\documentclass[10pt,journal]{IEEEtran}

\usepackage{cite}

\ifCLASSINFOpdf
  \usepackage[pdftex]{graphicx}
  \usepackage{epstopdf}
\else
\fi

\usepackage[cmex10]{amsmath}
\usepackage{amsfonts}

\usepackage{algorithm}
\usepackage{algorithmicx}
\usepackage{algpseudocode}

\usepackage[tight,footnotesize]{subfigure}

\usepackage{url}

\newtheorem{Assumption}{Assumption}
\newtheorem{Lemma}{Lemma}
\newtheorem{Theorem}{Theorem}
\newtheorem{Corollary}{Corollary}

\DeclareMathOperator*{\argmax}{\arg\!\max}

\usepackage{accents}

\newcommand{\ceil}[1]{\lceil{#1}\rceil}
 
\DeclareMathOperator{\ue}{ue}
\DeclareMathOperator{\mess}{message}
\DeclareMathOperator{\EXP}{``explore''}
\DeclareMathOperator{\explore}{or}
\DeclareMathOperator{\exploit}{oi}
\DeclareMathOperator{\all}{all}
\DeclareMathOperator{\LinUCB}{LinUCB}
\DeclareMathOperator{\AUER}{AUER}
\DeclareMathOperator{\PROPOSED}{HCL}
\DeclareMathOperator{\LC}{LC}
\DeclareMathOperator{\MCSP}{MCSP}
\DeclareMathOperator{\E}{E} 

\begin{document}

%
\title{Context-Aware Hierarchical Online Learning \\ for Performance Maximization \\ in Mobile Crowdsourcing}

\author{\IEEEauthorblockN{Sabrina Klos n\'{e}e M\"uller\IEEEauthorrefmark{1},
Cem Tekin\IEEEauthorrefmark{2},
Mihaela van der Schaar\IEEEauthorrefmark{3}$^{,}$\IEEEauthorrefmark{4} and
Anja Klein\IEEEauthorrefmark{1}}\\
\IEEEauthorblockA{\IEEEauthorrefmark{1}Communications Engineering Lab, TU Darmstadt, Germany, \{s.klos, a.klein\}@nt.tu-darmstadt.de}\\
\IEEEauthorblockA{\IEEEauthorrefmark{2}Electrical and Electronics Engineering Department, Bilkent University, Ankara, Turkey, cemtekin@ee.bilkent.edu.tr}\\
\IEEEauthorblockA{\IEEEauthorrefmark{3}Department of Electrical Engineering, University of California Los Angeles, USA, mihaela@ee.ucla.edu}\\
\IEEEauthorblockA{\IEEEauthorrefmark{4}Department of Engineering Science, University of Oxford, Oxford, UK} %
\thanks{This work is copyrighted by the IEEE. This version of the article has been accepted for publication in IEEE/ACM Transactions on Networking. 
The final published article can be found in IEEE’s electronic database under DOI: 10.1109/TNET.2018.2828415.}
}

\maketitle

\begin{abstract}
In mobile crowdsourcing (MCS), mobile users accomplish outsourced human intelligence tasks. 
MCS requires an appropriate task assignment strategy, since different workers may have different performance in terms of acceptance rate and quality. 
Task assignment is challenging, since a worker's performance (i) may fluctuate, depending on both the worker's current personal context and the task context, (ii) is not known a priori, but has to be learned over time. 
Moreover, learning context-specific worker performance requires access to context information, which may not be available at a central entity due to communication overhead or privacy concerns.
Additionally, evaluating worker performance might require costly quality assessments.
In this paper, we propose a context-aware hierarchical online learning algorithm addressing the problem of performance maximization in MCS. 
In our algorithm, a local controller (LC) in the mobile device of a worker regularly observes the worker's context, her/his decisions to accept or decline tasks and the quality in completing tasks. 
Based on these observations, the LC regularly estimates the worker's context-specific performance.
The mobile crowdsourcing platform (MCSP) then selects workers based on performance estimates received from the LCs.
This hierarchical approach enables the LCs to learn context-specific worker performance and it enables the MCSP to select suitable workers. 
In addition, our algorithm preserves worker context locally, and it keeps the number of required quality assessments low. 
We prove that our algorithm converges to the optimal task assignment strategy.
Moreover, the algorithm outperforms simpler task assignment strategies in experiments based on synthetic and real data.
\end{abstract}


\section{Introduction}
\textit{Crowdsourcing} (CS) is a popular way to outsource human intelligence tasks, prominent examples being conventional web-based systems like Amazon Mechanical Turk\footnote{https://www.mturk.com} and Crowdflower\footnote{https://www.crowdflower.com/}.
More recently, \textit{mobile crowdsourcing} (MCS) has evolved as a powerful tool to leverage the workforce of mobile users to accomplish tasks in a distributed manner~\cite{RenZhangZhangEtAl2015}. 
This may be due to the fact that the number of mobile devices is growing rapidly and at the same time, people spend a considerable amount of their daily time using these devices.
For example, between 2015 and 2016, the global number of mobile devices grew from 7.6~to~8~billion~\cite{cisco}. 
Moreover, the daily time US adults spend using mobile devices is estimated to be more than 3 hours in 2017, which is an increase by more than $45\%$ compared to 2013~\cite{emarketer}.

In MCS, \textit{task owners} outsource their tasks via an intermediary \textit{mobile crowdsourcing platform} (MCSP) to a set of \textit{workers}, i.e.,  mobile users, who may complete these tasks.
An MCS task may require the worker to interact with her/his mobile device in the physical world (e.g., photography tasks) or to complete some virtual task via the mobile device (e.g., image annotation, sentiment analysis).
Some MCS tasks, subsumed under the term \textit{spatial} CS~\cite{ZhaoHan2016}, are spatially constrained (e.g., photography task at point of interest), or require high spatial resolution (e.g., air pollution map of a city).
In spatial CS, tasks typically require workers to travel to certain locations.  
However, recently emerging MCS applications are also concerned with \textit{location-independent} tasks. 
For example, MapSwipe\footnote{https://mapswipe.org/} lets mobile users annotate satellite imagery to find inhabitated regions around the world.
The GalaxyZoo app\footnote{https://www.galaxyzoo.org/} lets mobile users classify galaxies. 
The latter project is an example of the more general trend of citizen science~\cite{BonneyShirkPhillipsEtAl2014}.
On the commercial side, Spare5\footnote{https://app.spare5.com/fives} or Crowdee\footnote{https://www.crowdee.de/} outsource micro-tasks (e.g., image annotation, sentiment analysis, and opinion polls) to mobile users in return for small payments.
While location-independent tasks could as well be completed by users of static devices as in web-based CS, emerging MCS applications for location-independent tasks exploit that online mobile users complete such tasks on the go.

MCS -- be it spatial or location-independent -- requires an appropriate \textit{task assignment} strategy, since not all workers may be equally suitable for a given task. 
First, different workers may have different task preferences and hence different acceptance rates. 
Secondly, different workers may have different skills, and hence provide different quality when completing a task.
Two assignment modes considered in the CS literature are the \textit{server assigned tasks} (SAT) mode and the \textit{worker selected tasks} (WST) mode~\cite{KazemiShahabi2012}. 
In SAT mode, the MCSP tries to match workers and tasks in an optimal way, e.g., to maximize the number of task assignments, possibly under a given task budget. 
For this purpose, the MCSP typically gathers task and worker information to decide on task assignment. 
This sophisticated strategy may require a large communication overhead and a privacy concern for workers since the MCSP has to be regularly informed about the current worker contexts (e.g., their current positions).
Moreover, previous work on the SAT mode often either assumed that workers always accept a task once assigned to it or that workers' acceptance rates and quality are known in advance. 
However, in reality, acceptance rates and quality are usually not known beforehand and therefore have to be learned by the MCSP.
In addition, a worker's acceptance rate and the quality of completed tasks might depend not only on the specific task, but also on the worker's current context, e.g., the worker's location or the time of day~\cite{GeigerSchader2014}. 
This context may change quickly, especially in MCS with location-independent tasks, since workers can complete such tasks anytime and anywhere.

In contrast, in~WST mode, workers autonomously select tasks from a list. 
This rather simple mode is often used in practice (e.g., on Amazon Mechanical Turk) since it has the advantage that workers automatically select tasks they are interested in. 
However, the WST mode can lead to suboptimal task assignments since first, finding suitable tasks is not as easy as it seems (e.g., time-consuming searches within a long list of tasks are needed and workers might simply select from the first displayed tasks~\cite{ChiltonHortonMillerEtAl2010}) and secondly, workers might leave unpopular tasks unassigned.
Therefore, in WST mode, the MCSP might additionally provide personalized \textit{task recommendation} (TR) to workers such that workers find appropriate tasks~\cite{GeigerSchader2014}. 
However, personalized TR typically requires workers to share their current context with the MCSP, which again may mean a communication overhead and a privacy concern for workers.

We argue that a task assignment strategy is needed which combines the advantages of the above modes:
The MCSP should centrally coordinate task assignment to ensure that appropriate workers are selected, as in SAT mode.
At the same time, the workers' personal contexts should be kept locally, as in WST mode, in order to keep the communication overhead small and to protect the workers' privacy.
Moreover, task assignment should take into account that workers may decline tasks, and hence, the assignment should fit to the workers' preferences, as in WST mode with personalized~TR.
In addition, task assignment should be based both on acceptance rates and on the quality with which a task is completed.
Since quality assessments (e.g., a manual quality rating from a task owner, or an automatic quality assessment using either local software in a mobile device or the resources of a cloud) may be costly, the number of quality assessments should be kept low.
Finally, workers' acceptance rates and quality have to be learned over time.

Our contribution therefore is as follows: We propose a context-aware hierarchical online learning algorithm for performance maximization in MCS for location-independent tasks. 
Our algorithm for the first time \textit{jointly} takes the following aspects into account: $(i)$~Our algorithm learns worker performance online without requiring a training phase. Since our algorithm learns in an online fashion, it adapts and improves the worker selection over time and can hence achieve good results already during run time. By establishing regret bounds, we provide performance guarantees for the learned task assignment strategy and prove that our algorithm converges to the optimal task assignment strategy. $(ii)$~We allow different task types to occur. We use the concept of \textit{task context} to describe the features of a task, such as its required skills or equipment. $(iii)$~We model that the worker performance depends (in a possibly non-linear fashion) on both the task context and the \textit{worker context}, such as the worker's current location, activity, or device status. Our proposed algorithm learns this context-specific worker performance.  $(iv)$~Our algorithm is split into two parts, one part executed by the MCSP, the other part by \textit{local controllers} (LCs) located in each of the workers' mobile devices.
An LC learns its worker's performance in terms of acceptance rate and quality online over time, by observing the worker's personal contexts, her/his decisions to accept or decline tasks and the quality in completing these tasks. 
The LC learns from its worker's context only locally, and personal context is not shared with the MCSP. 
Each LC regularly sends performance estimates to the MCSP. 
Based on these estimates, the MCSP takes care of the worker selection. 
This hierarchical (in the sense of the coordination between the MCSP and the LCs) approach enables the MCSP to select suitable workers for each task under its budget based on what the LCs have previously learned.
Moreover, workers receive personalized task requests based on their interests and skills, while keeping the number of (possibly costly) quality assessments low.

The remainder of this paper is organized as follows. 
Sec.~\ref{Sec_rel_work} gives an overview on related work.
Sec.~\ref{Sec_system_model} describes the system model. 
In Sec.~\ref{Sec_Algo}, we propose a context-aware hierarchical online learning algorithm for performance maximization in MCS. 
In Sec.~\ref{Sec_analysis}, we theoretically analyze our algorithm in terms of its regret, as well as its requirements with respect to local storage, communication and worker quality assessment. 
Sec.~\ref{Sec_num_results} contains a numerical evaluation based on synthetic and real data.
Sec.~\ref{Sec_conclusion} concludes the paper.

\section{Related Work}\label{Sec_rel_work}

Research has put some effort in theoretically defining and classifying CS systems, such as web-based~\cite{DoanRamakrishnanHalevy2011}, mobile~\cite{RenZhangZhangEtAl2015} and spatial~\cite{ZhaoHan2016} CS.
\nocite{AhujaSchaar2016}
\nocite{LiuLiu2017}
\nocite{SlivkinsVaughan2013}
\begin{table*}[!t]
\renewcommand{\arraystretch}{1.1}
\renewcommand{\tabcolsep}{1.8mm}
\caption{Comparison with related work on task assignment in crowdsourcing.}
\label{Table_related_works}
   \centering
      \begin{tabular}{|c|c|c|c|c|c|c|c|c|c|c|c|}
      \hline
 			& \cite{HoVaughan2012}  & \cite{Tran-ThanhSteinRogersEtAl2014} & \cite{SafranChe2017} & \cite{AmbatiVogelCarbonell2011} & \cite{GongWeiGuoEtAl2016} & \cite{HanZhangLuo2016} & \cite{KazemiShahabi2012},\cite{ToShahabiKazemi2015}  & \cite{HassanCurry2014} & \cite{ToGhinitaFanEtAl2016} & \cite{ZhengChen2017}  & This work\\
       \hline
      Crowdsourcing Type & General & General & General & General& Mobile & Mobile & Spatial & Spatial & Spatial & Spatial & Loc.-Ind. \\
      \hline
      Task Assignment Mode & SAT & SAT & SAT & WST/TR & WST/TR & SAT & SAT & SAT & SAT & SAT & proposed\\
      \hline
      Different Task Types & Yes  & No & Yes & Yes & Yes & No & Yes & Yes & No & Yes & Yes\\
      \hline
      Worker Context-Aware & No  & No & No & No & Yes & No & Yes & Yes & Yes & No & Yes\\
      \hline
      Context-Spec. Performance  & No  & No & No & No & Yes & No & No & Yes & Yes & No & Yes\\
      \hline
      Worker Context Protected & N/A & N/A & N/A & N/A & Yes & N/A & No & No & Yes & N/A & Yes\\
      \hline
      Type of Learning & Online & Online & Batch & Online & Offline & Online & N/A & Online & N/A & Offline & Online\\
      \hline
      Regret Bounds & Yes  & Yes & No & No & No & Yes & N/A & Yes & N/A & No & Yes\\
      \hline
      \end{tabular}
\end{table*}
Below, we give an overview on related work on task assignment in general, mobile and spatial CS systems, as relevant for our scenario.
Note that strategic behavior of workers and task owners in CS systems, e.g., concerning pricing and effort spent in task completion~\cite{AhujaSchaar2016} is out of the scope of this paper.
Also note that we assume that it is possible to assess the quality of a completed task. A different line of work on CS deals with quality estimation in case of missing ground truth, recently also using online learning~\cite{LiuLiu2017}. 

Due to the dynamic nature of CS, with tasks and/or workers typically arriving dynamically over time, task assignment is often modeled as an online decision making problem~\cite{SlivkinsVaughan2013}. 
For general CS systems, \cite{HoVaughan2012} proposed a competitive online task assignment algorithm for maximizing the utility of a task owner on a given set of task types, with finite number of tasks per type, by learning the skills of sequentially appearing workers.
While~\cite{HoVaughan2012} considers sequentially arriving workers and their algorithm decides which task to assign to a worker, we consider sequentially arriving tasks and our algorithm decides which workers to assign to a task. 
Therefore, our algorithm can be applied to an infinite number of task types by describing a task using its context. 
In addition, our algorithm takes worker context into account, which may affect worker performance in MCS.
In~\cite{Tran-ThanhSteinRogersEtAl2014}, a bounded multi-armed bandit model for expert CS is presented and a task assignment algorithm with sublinear regret is derived which maximizes the utility of a budget-constrained task owner under uncertainty about the skills of a finite set of workers with (known) different prices and limited working time.
While in~\cite{Tran-ThanhSteinRogersEtAl2014}, the average skill of a worker is learned, our algorithm takes context into account, and thereby learns context-specific performance.
In~\cite{SafranChe2017}, a real-time algorithm for finding the top-k workers for sequentially arriving tasks is presented. 
First, tasks are categorized offline into different types and the similarity between a worker's profile and each task type is computed. 
Then, in real time, the top-k workers are selected for a task based on a matching score, which takes into account the similarity and historic worker performance. 
The authors propose to periodically update the performance estimates offline in batches, but no guarantees on the learning process are given. 
In contrast, we additionally take into account worker context, learn context-specific performance and derive guarantees on the learning speed.
In~\cite{AmbatiVogelCarbonell2011}, methods for learning a worker preference model are proposed for personalized TR in WST mode. 
These methods use the history of worker preferences for different tasks, but they do not take into account worker context.

For MCS systems,~\cite{GongWeiGuoEtAl2016} proposes algorithms for optimal TR in WST mode that take into account the trade-off between the privacy of worker context, the utility to recommend the best tasks and the efficiency in terms of communication and computation overhead. 
TR is performed by a server based on a generalized context shared by the worker.
The statistics used for TR are gathered offline via a proxy that ensures differential privacy guarantees. 
While~\cite{GongWeiGuoEtAl2016} allows to flexibly adjust the shared generalized context and makes TRs based on offline statistics and generalized worker context, our approach keeps worker context locally and learns each worker's individual statistics online. 
In~\cite{HanZhangLuo2016}, an online learning algorithm for mobile crowdsensing is presented to maximize the revenue of a budget-constrained task owner by learning the sensing values of workers with known prices. 
While~\cite{HanZhangLuo2016} considers a total budget and each crowdsensing task requires a minimum number of workers, we consider a separate budget per task, which translates to a maximum number of required workers, and we additionally take task and worker context into account.

A taxonomy for spatial CS was first introduced in \cite{KazemiShahabi2012}.
The authors present a location-entropy based algorithm for SAT mode to maximize the number of task assignments under uncertainty about task and worker arrival processes. 
The server decides on task assignment based on centrally gathered knowledge about the workers' current locations. 
In~\cite{ToShahabiKazemi2015}, the authors extend this framework to maximize the quality of assignments under varying worker skills for different task types.
However, in contrast to our work,~\cite{KazemiShahabi2012}~and~\cite{ToShahabiKazemi2015} assume that worker context is centrally gathered, that workers always accept assigned tasks within certain known bounds and that worker skills are known a priori. 
In~\cite{HassanCurry2014}, an online task assignment algorithm is proposed for spatial CS with SAT mode for maximizing the expected number of accepted tasks.
The problem is modeled as a contextual multi-armed bandit problem, and workers are selected for sequentially arriving tasks. 
The authors adapt the LinUCB algorithm by assuming that the acceptance rate is a linear function of the worker's distance to the task and the task type. 
However, such a linearity assumption is restrictive and it especially may not hold in MCS with location-independent tasks. 
In contrast, our algorithm works for more general relationships between context and performance. 
In~\cite{ToGhinitaFanEtAl2016}, an algorithm for privacy-preserving spatial CS in SAT mode is proposed. 
Using differential privacy and geocasting, the algorithm preserves worker locations (i.e., their contexts) while optimizing the expected number of accepted tasks. 
However, the authors assume that the workers' acceptance rates are identical and known, whereas our algorithm learns context-specific acceptance rates.
In~\cite{ZhengChen2017}, exact and approximation algorithms for acceptance maximization in spatial CS with SAT mode are proposed. 
The algorithms are performed offline for given sets of available workers and tasks based on a probability of interest for each pair of worker and task. 
The probabilities of interest are computed beforehand using maximum likelihood estimation. 
On the contrary, our algorithm learns acceptance rates online and we provide an upper bound on the regret of this learning.

We model the problem formally as a \textit{contextual multi-armed bandit} (contextual MAB) problem~\cite{LangfordZhang2007, BadanidiyuruLangfordSlivkins2014, LiChuLangfordEtAl2010, ChuLiReyzinEtAl2011, AgrawalGoyal2013, GentileLiZappella2014, Slivkins2014, Tekin.Schaar2015, TekinZhangSchaar2014, MuellerAtanSchaarEtAl2016, MuellerAtanSchaarEtAl2016b}. 
MABs are a type of \textit{reinforcement learning} (RL). 
In general, RL, which has been used to solve various problems in networking~\cite{ShiangSchaar2008, MastronardeSchaar2011}, deals with agents learning to take actions based on rewards.
Specifically, in contextual MAB, an agent sequentially chooses among a set of actions with unknown expected rewards. In each round, the agent first observes some context information, which he may use to determine the action to select. After selecting an action, the agent receives a reward, which may depend on the context. The agent tries to learn which action has the highest reward in which context, to maximize his expected reward over time.

In the related literature on contextual MAB, different algorithms make different assumptions on how context is generated and on how rewards are formed.  
For general contextual MAB with no further assumptions on how rewards are formed, Ref.~\cite{LangfordZhang2007} proposes the epoch-greedy algorithm. 
Also Ref. \cite{BadanidiyuruLangfordSlivkins2014} for general contextual MAB with resource constraints and policy sets makes no further assumptions on how rewards are formed, except that they assume that the marginal distribution over the contexts is known.
However, the algorithms in \cite{LangfordZhang2007},\cite{BadanidiyuruLangfordSlivkins2014} work only for a finite set of actions and they assume that at each time step the tuples (context, rewards) are sampled from a fixed but unknown distribution (i.e., contexts are generated in an i.i.d. fashion). 
Other algorithms have stronger assumptions on how rewards are formed. For example, the LinUCB algorithm~\cite{LiChuLangfordEtAl2010},~\cite{ChuLiReyzinEtAl2011}, assumes that the expected reward is linear in the context. 
Such a linearity assumption is also used in the Thompson-sampling based algorithm in \cite{AgrawalGoyal2013}, and in the clustering algorithm in \cite{GentileLiZappella2014}, where a clustering is performed on top of a contextual MAB setting.
There are also works which assume a known similarity metric over the contexts. 
These algorithms group the contexts into sets of similar contexts by partitioning the context space. 
Then, they estimate the reward of an action under a given context based on previous rewards for that action in the set of similar contexts. 
For example, the contextual zooming algorithm~\cite{Slivkins2014} proposes a non-uniform adaptive partition of the context space. 
Moreover,~\cite{Tekin.Schaar2015, TekinZhangSchaar2014} use uniform and non-uniform adaptive partitions of the context space. In~\cite{MuellerAtanSchaarEtAl2016, MuellerAtanSchaarEtAl2016b}, these algorithms are applied to a wireless communication scenario. 
While the algorithms in \cite{LiChuLangfordEtAl2010, ChuLiReyzinEtAl2011,AgrawalGoyal2013,
GentileLiZappella2014,Slivkins2014, Tekin.Schaar2015, TekinZhangSchaar2014, MuellerAtanSchaarEtAl2016, MuellerAtanSchaarEtAl2016b} are more restrictive with respect to how rewards are formed, they are more general than  \cite{LangfordZhang2007},\cite{BadanidiyuruLangfordSlivkins2014} in the sense that they do not require the contexts to be generated i.i.d. over time. Moreover, the algorithms in \cite{Slivkins2014, Tekin.Schaar2015, TekinZhangSchaar2014, MuellerAtanSchaarEtAl2016, MuellerAtanSchaarEtAl2016b} also work for an infinite set of actions.

Algorithms for contextual MAB also differ with respect to their approach to balance the exploration vs. exploitation trade-off. While the epoch-greedy algorithm~\cite{LangfordZhang2007} and the algorithms in~\cite{Tekin.Schaar2015, TekinZhangSchaar2014, MuellerAtanSchaarEtAl2016, MuellerAtanSchaarEtAl2016b} explicitly distinguish between exploration and exploitation steps, the LinUCB~\cite{LiChuLangfordEtAl2010},~\cite{ChuLiReyzinEtAl2011} algorithm, the clustering algorithm in~\cite{GentileLiZappella2014} and the contextual zooming algorithm~\cite{Slivkins2014} follow an index-based approach, in which in any round, the action with the highest index is selected. Other algorithms, like the one for contextual MAB with resource constraints in Ref. \cite{BadanidiyuruLangfordSlivkins2014}, draw samples from a distribution to find a policy which is then used to select the action. 
Finally, algorithms like the Thompson-sampling based algorithm in \cite{AgrawalGoyal2013} draw samples from a distribution to build a belief, and select the action which maximizes the expected reward based on this belief.

Our proposed algorithm extends~\cite{Tekin.Schaar2015, TekinZhangSchaar2014, MuellerAtanSchaarEtAl2016, MuellerAtanSchaarEtAl2016b} as follows: $(i)$~While in~\cite{Tekin.Schaar2015, TekinZhangSchaar2014, MuellerAtanSchaarEtAl2016, MuellerAtanSchaarEtAl2016b}, a learner observes some contexts and selects a subset of actions based on these contexts, our algorithm is decoupled to several learning entities, each observing the context of one particular action and learning the rewards of this action, and a coordinating entity, which selects a subset of actions based on the learning entities' estimates. In the MCS scenario, an action corresponds to a worker, the learning entities correspond to the LCs which learn the performance of their workers, and the coordinating entity corresponds to the MCSP, which selects workers based on the performance estimates from the LCs.
$(ii)$~While in~\cite{Tekin.Schaar2015, TekinZhangSchaar2014, MuellerAtanSchaarEtAl2016, MuellerAtanSchaarEtAl2016b}, the same number of actions is selected per round, we allow different numbers of actions to be selected per round. In the MCS scenario, this corresponds to allowing different numbers of required workers for different tasks. Hence, in contrast to~\cite{Tekin.Schaar2015, TekinZhangSchaar2014, MuellerAtanSchaarEtAl2016, MuellerAtanSchaarEtAl2016b}, the learning speed of our algorithm is affected by the arrival process of the numbers of actions to be selected.
$(iii)$~While in~\cite{Tekin.Schaar2015, TekinZhangSchaar2014, MuellerAtanSchaarEtAl2016, MuellerAtanSchaarEtAl2016b}, each action has the same context space, we allow each action to have an individual context space of an individual dimension. In the MCS scenario, this corresponds to allowing workers to give access to individual sets of context dimensions. Therefore, in contrast to~\cite{Tekin.Schaar2015, TekinZhangSchaar2014, MuellerAtanSchaarEtAl2016, MuellerAtanSchaarEtAl2016b}, the granularity of learning may be different for different actions. 
$(iv)$~Finally, while in~\cite{Tekin.Schaar2015, TekinZhangSchaar2014, MuellerAtanSchaarEtAl2016, MuellerAtanSchaarEtAl2016b}, all actions are available in any round, we allow actions to be unavailable in arbitrary rounds. In the MCS scenario, this corresponds to allowing that workers may be unavailable. Hence, in contrast to~\cite{Tekin.Schaar2015, TekinZhangSchaar2014, MuellerAtanSchaarEtAl2016, MuellerAtanSchaarEtAl2016b}, the best subset of actions in a certain round depends on the specific set of available actions in this round.

\section{System Model}\label{Sec_system_model}
\subsection{Mobile Crowdsourcing Platform}
We consider an MCSP, to which a fixed set $\mathcal{W}$ of $W:=|\mathcal{W}|$ workers belongs.
A worker is a user equipped with a mobile device, in which the MCS application is installed. 
Workers can be in two modes: 
A worker is called \textit{available}, if the MCS application on the device is running. 
In this case, the MCSP may request the worker to complete a task, which the worker may then accept or decline.
A worker is called \textit{unavailable}, if the MCS application on the device is turned off.

Task owners can place location-independent tasks of different types into the MCSP and select a task budget. 
A task $t$ is defined by a tuple $(b_t, c_t)$, where $b_t> 0$ denotes the budget that the task owner is willing to pay for this task and $c_t\in \mathcal{C}$ denotes the task context. 
The task context is taken from a bounded $C$-dimensional task context space $\mathcal{C}:=[0,1]^{C}$ and captures feature information about the task.\footnote{We assume that tasks are described by $C$ context dimensions. In each of the $C$ context dimensions, a task is classified via a value between $[0,1]$. Then, $c_t\in[0,1]^C$ is a vector describing task $t$'s overall context.} Possible features could be the skills or equipment required to complete a task (e.g., the required levels of creativity or analytical skills may be translated to continuous values between $0$ and $1$; whether a camera or a specific application is needed may be encoded as $0$ or $1$).
The task owner has to pay the MCSP for each worker that completes the task after being requested by the MCSP. Specifically, we assume that the MCSP charges the task owner a fixed price $e_t\in[e_{\min}, e_{\max}]$ per worker that completes task~$t$, where $e_{\min}>0$ and $e_{\max}\geq e_{\min}$ correspond to lower and upper price limits, respectively. The price $e_t$ depends on the task context $c_t$ and is determined by the MCSP's fixed context-specific price list. 
We assume that for each task $t$, the budget $b_t$ satisfies $b_t\in [e_t,We_t]$, so that the budget is sufficient to pay at least one and at most $W$ workers for completing the task.
Based on the budget $b_t$ and the price $e_t$, the MCSP computes the maximum number $m_t:=\lfloor \frac{b_t}{e_t} \rfloor \in \{1,...,W\}$ of workers that should complete the task. 

Following~\cite{HoVaughan2012, Tran-ThanhSteinRogersEtAl2014, HanZhangLuo2016}, we assume that each task has the following properties: $(i)$~As determined by budget and price, the task owner would like to receive replies from possibly several workers. $(ii)$~It is possible to assess the quality of a single worker's reply. $(iii)$~The qualities of different workers' replies are independent. $(iv)$~The qualities of the workers' replies are additive, i.e., if workers $1$ and $2$  complete the task with qualities $A$ and $B$, the task owner receives a total quality of $A+B$. Such tasks belong to the class of \textit{crowd solving} tasks~\cite{GeigerSchader2014}, examples being translation and retrieval tasks~\cite{HoVaughan2012}. 

We assume that tasks arrive at the MCSP sequentially
and we denote the sequentially arriving tasks by $t=1,...,T$. 
For each arriving task~$t$, if sufficient workers are available, the MCSP will request $m_t$ workers to complete the task.\footnote{Note that each task is only processed once by the MCSP, even if not all $m_t$ requested workers complete the task. In this case, the MCSP charges the task owner only for the actual number of workers that completed the task since only these workers are compensated. The task owner may submit the task to the MCSP again if she/he wishes more workers to complete the task.}
\begin{figure}[!t]
\centering
\includegraphics[width=0.46\textwidth]{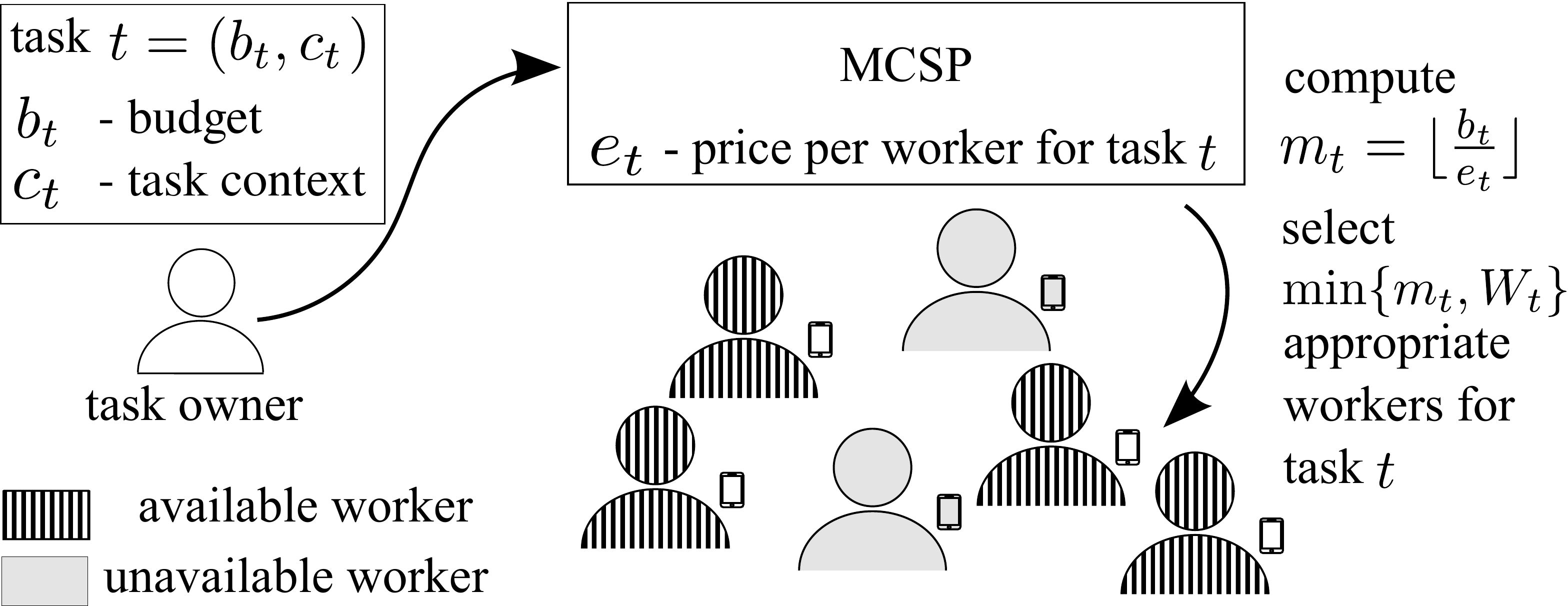}
\caption{System model. A task arrives at the MCSP. The MCSP has to select an appropriate subset of available workers for the task.}
\label{Fig_system_model}
\end{figure}
However, due to the dynamics in worker availability over time, the MCSP can only select workers from the set $\mathcal{W}_t\subseteq \mathcal{W}$ of currently available workers for task $t$, as defined by $\mathcal{W}_t:= \{i: \text{ worker } i \text{ is available at arrival time of } t\}$, where the number of available workers\footnote{We assume that for each arriving task, at least one worker is available.} is denoted by $W_t := |\mathcal{W}_t|\in \{1,...,W\}$. 
Hence, since the MCSP can select at most all available workers, it aims at selecting $\min\{m_t,W_t\}$ workers for task $t$.\footnote{If fewer than $m_t$ workers are available, the MCSP will request all available workers to complete the task.} The goal of the MCSP is to select a subset of $\min\{m_t,W_t\}$ workers which maximizes the worker performance for that task (see Fig.~\ref{Fig_system_model} for an illustration).

\subsection{Context-Specific Worker Performance}
The performance of a worker depends on $(i)$~the worker's willingness to accept the task and $(ii)$~the worker's quality in completing the task, where we assume that the quality can take values in a range $[q_{\min},q_{\max}]\subseteq \mathbb{R}_{0,+}$. 
Both the willingness to accept the task and the quality may depend on the worker's current context and on the task context.
Let $x_{t,i}$ denote the personal context of worker $i\in \mathcal{W}_t$ at the arrival time of task~$t$, coming from a bounded $X_i$-dimensional personal context space~$\mathcal{X}_i:=[0,1]^{X_i}$.
Here, we allow each worker $i$ to have an individual personal context space $\mathcal{X}_i$, since each worker may allow access to an individual set of context dimensions (e.g., the worker allows access to a certain set of sensors of the mobile device that are used to derive her/his context).
Possible personal context dimensions could be the worker's current location (in terms of geographic coordinates), the type of location (e.g., at home, in a coffee shop), the worker's current activity (e.g., commuting, working) or the current device status (e.g., battery state, type of wireless connection).
We further call the concatenation $(x_{t,i},c_t)\in \mathcal{X}_i \times \mathcal{C}$ the \textit{joint (personal and task) context of worker $i$}. 
For worker $i$, this joint context is hence a vector of dimension $D_i:=X_i+C$. 
We call  $\mathcal{X}_i \times \mathcal{C} = [0,1]^{X_i}\times[0,1]^{C} \equiv [0,1]^{D_i}$  the \textit{joint (personal and task) context space of worker $i$}.
The reason for considering the joint context is that the performance of worker $i$ may depend on both the current context $x_{t,i}$ and the task context $c_t$ -- in other words, the performance depends jointly on $(x_{t,i},c_t)$.

Let $p_i(x_{t,i},c_t)$ denote the performance of worker $i$ with current personal context $x_{t,i}$ for task context $c_t$. 
The performance can be decomposed into $(i)$~worker $i$'s decision $d_i(x_{t,i},c_t)$ to accept ($d_i(x_{t,i},c_t)=1$) or reject ($d_i(x_{t,i},c_t)=0$) the task and, in case the worker accepts the task, also on $(ii)$~worker $i$'s quality $q_i(x_{t,i},c_t)$ when completing the task. 
Hence, we can write
\begin{align*}
p_i(x_{t,i},c_t)&:=\begin{cases}
q_i(x_{t,i},c_t), \ \ &\text{ if } d_i(x_{t,i},c_t)=1,\\
0, \ \ &\text{ if } d_i(x_{t,i},c_t)=0.
\end{cases}
\end{align*}
The performance is a random variable whose distribution depends on the distributions of the random variables $d_i(x_{t,i},c_t)$ and $q_i(x_{t,i},c_t)$. 
Since the decision $d_i(x_{t,i},c_t)$ is binary, it is drawn from the Bernoulli distribution with unknown parameter $r_i(x_{t,i},c_t) \in [0,1]$.
Hence, $r_i(x_{t,i},c_t)$ represents worker $i$'s acceptance rate given the joint context $(x_{t,i},c_t)$.
The quality $q_i(x_{t,i},c_t)$ is a random variable conditioned on $d_i(x_{t,i},c_t)=1$ (i.e., task acceptance) with unknown distribution and we denote its expected value by $\nu_i(x_{t,i},c_t):=\E[q_i(x_{t,i},c_t)]$. 
Hence, $\nu_i(x_{t,i},c_t)$ represents the average quality of worker~$i$ with personal context $x_{t,i}$ when completing a task of context~$c_t$.
Therefore, the performance $p_i(x_{t,i},c_t)$ of worker $i$ given the joint context $(x_{t,i},c_t)$ has unknown distribution, takes values in $[0,q_{\max}]$ and its expected value satisfies
\begin{align}
\E[p_i(x_{t,i},c_t)]& = \theta_i(x_{t,i},c_t), \notag 
\end{align}
where $\theta_i(x_{t,i},c_t) := r_i(x_{t,i},c_t)\nu_i(x_{t,i},c_t)$. 

\subsection{Problem Formulation}\label{Sec_Problem_Formulation}
Consider an arbitrary sequence of task and worker arrivals.\footnote{In the following, by ``\textit{an arbitrary sequence of task and worker arrivals}'', we mean, given arbitrary sequences of task budgets $\{b_t\}_{t=1,...,T}$, task contexts $\{c_t\}_{t=1,...,T}$, task prices $\{e_t\}_{t=1,...,T}$, worker availability ${\{\mathcal{W}_t\}}_{t=1,...,T}$ and worker contexts $\{x_{t,i}\}_{i\in \mathcal{W}_t,t=1,...,T}$.}
Let $y_{t,i}$ denote a binary variable which is $1$ if worker $i$ is requested to complete task $t$ and $0$ otherwise. 
Then, the problem of selecting, for each task, a subset of workers which maximizes the sum of expected performances given the task budget is given by
\begin{align}\label{Eq_Problem}
&\max_{\{y_{t,i}\}_{i\in \mathcal{W}_t, t=1,...,T}} &&\sum_{t=1}^T\sum_{i\in \mathcal{W}_t} \theta_i(x_{t,i},c_t) y_{t,i} \\
&\text{s.t. } &&\sum_{i\in \mathcal{W}_t} y_{t,i} \leq m_t \ \forall t=1,...,T\notag\\
&&&y_{t,i}\in\{0,1\} \ \ \forall i\in \mathcal{W}_t, \  \forall t=1,...,T.\notag
\end{align}
First, we analyze problem~\eqref{Eq_Problem} assuming full knowledge about worker performance. Therefore, assume that there was an entity that $(i)$~was an omniscient \textit{oracle}, which knows the expected performance of each worker in each context for each task context \textit{a priori} and $(ii)$~for each arriving task, this entity is \textit{centrally} informed about the current contexts of all available workers. 
For such an entity, problem~\eqref{Eq_Problem} is an integer linear programming problem,
which can be decoupled to an independent sub-problem per arriving task.
For a task~$t$, if fewer workers are available than required, i.e., $W_t\leq m_t$, the optimal solution is to request all available workers to complete the task.
However, if $W_t > m_t$, the corresponding sub-problem is a special case of a knapsack problem with a knapsack of size $m_t$ and with items of identical size and non-negative profit. 
Therefore, the optimal solution can be easily computed in at most $O(W\log(W))$ by ranking the available workers according to their context-specific expected performance and selecting the $m_t$ highest ranked workers.
By $\mathcal{S}_t^{*}:=\{s_{t,1}^{*},...,s_{t,\min\{m_t,W_t\}}^{*}\}$, we denote the optimal subset of workers to select for task $t$. 
Formally, these workers satisfy
\begin{align*}
s_{t,j}^{*}\in \argmax_{i\in\mathcal{W}_t\setminus \bigcup_{k=1}^{j-1}\{s_{t,k}^{*}\}}\hspace{-0.2cm}\theta_i(x_{t,i},c_t) \ \text{ for } j=1,...,\min\{m_t,W_t\},
\end{align*}
where $\bigcup_{k=1}^{0}\{s_{t,k}^{*}\}:=\emptyset$. Note that $\mathcal{S}_t^{*}$ depends on the task budget $b_t$, context $c_t$, price $e_t$, the set $\mathcal{W}_t$ of available workers and their personal contexts $\{x_{t,i}\}_{i\in \mathcal{W}_t}$, but we write $\mathcal{S}_t^{*}$ instead of $\mathcal{S}_t^{*}(b_t,c_t,e_t,\mathcal{W}_t, \{x_{t,i}\}_{i\in \mathcal{W}_t})$ for brevity. 
Let $\mathcal{S}^{*}:=\{\mathcal{S}_t^{*}\}_{t=1,...,T}$ be the collection of optimal subsets of workers for the collection $\{1,...,T\}$ of tasks. 
We call this collection the solution achieved by a \textit{centralized oracle}, since it requires an entity with a priori knowledge about expected performances and with access to worker contexts to make optimal decisions.

However, we assume that the MCSP does not have a priori knowledge about expected performances, but it still has to select workers for arriving tasks. 
Let $\mathcal{S}_t := \{s_{t,1},...,s_{t,\min\{m_t,W_t\}}\}$ denote the set of workers that the MCSP selects and requests to complete task~$t$. 
If for an arriving task, fewer workers are available than required, i.e., $W_t\leq m_t$, by simply requesting all available workers (i.e., $\mathcal{S}_t = \mathcal{W}_t$) to complete the task, the MCSP automatically selects the optimal subset of workers.
Otherwise, for $W_t> m_t$, the MCSP cannot simply solve problem~\eqref{Eq_Problem} like an omniscient oracle, since it does not know the expected performances $\theta_i(x_{t,i},c_t)$.
Moreover, we assume that a worker's current personal context is only locally available in the mobile device. 
We call the software of the MCS application, which is installed in the mobile device, a \textit{local controller} (LC) and we denote by LC~$i$ the LC of worker~$i$.
Depending on the requirements of the specific MCS application, such as, concerning communication overhead and worker privacy, the LCs may be owned by either the MCSP, the workers, or a trusted third party~\cite{GongWeiGuoEtAl2016, ToGhinitaFanEtAl2016}.
In any case, each LC has access to its corresponding worker's personal context, but it does not share this information with the MCSP. 

Hence, the MCSP and the LCs should cooperate in order to learn expected performances over time and in order to select an appropriate subset of workers for each task. 
For this purpose, over time, the system of MCSP and LCs has to find a trade-off between exploration and exploitation, by, on the one hand, selecting workers about whose performance only little information is available and, on the other hand, selecting workers which are likely to have high performance.
For each arriving task, the selection of workers depends on the history of previously selected workers and their observed performances. 
However, observing worker performance requires quality assessments (e.g., in form of a manual quality rating from a task owner, or an automatic quality assessment using either local software in the battery-constrained mobile device or the resources of a cloud), which may be costly.
Our model and algorithm are agnostic to the specific type of quality assessment, as long as the LCs do have access to the quality assessments. In any case, we aim at limiting the number of performance observations in order to keep the cost for quality assessment low. 

Next, we present a \textit{context-aware hierarchical online learning algorithm}, which maps the history of previously selected workers and observed performances to the next selection of workers.
The performance of this algorithm can be evaluated by comparing its loss with respect to the centralized oracle. 
This loss is called the \textit{regret of learning}.
For an arbitrary sequence of task and worker arrivals, 
the regret is formally defined as 
\begin{align}\label{Eq_regret}
& R(T) = \E\left[\sum_{t=1}^T \hspace{-0.1cm} \sum_{j=1}^{\min\{m_t,W_t\}} \hspace{-0.2in} \left( p_{s_{t,j}^{*}}(x_{t,s_{t,j}^{*}},c_t)
-p_{s_{t,j}}(x_{t,s_{t,j}},c_t)\right) \hspace{-0.1cm} \right]\hspace{-0.1cm},
\end{align}
which is equivalent to 
\begin{align}
R(T) = \sum_{t=1}^T \hspace{-0.1cm} \sum_{j=1}^{\min\{m_t,W_t\}}\hspace{-0.2cm}\left( \theta_{s_{t,j}^{*}}(x_{t,s_{t,j}^{*}},c_t) 
- \text{E}[ \theta_{s_{t,j}}(x_{t,s_{t,j}},c_t) ] \right)\hspace{-0.1cm}.
\end{align}
Here, the expectation is taken with respect to the selections made by the learning algorithm and the randomness of the workers' performances.

\section{A Context-aware Hierarchical Online Learning Algorithm for Performance Maximization in Mobile Crowdsourcing}\label{Sec_Algo}

The goal of the MCSP is to select, for each arriving task, a set of workers that maximizes the sum of expected performances for that task given the task budget.
Since the expected performances are not known a priori by neither MCSP nor the LCs, they have to be learned over time.
Moreover, since only the LCs have access to the personal worker contexts, a coordination is needed between the MCSP and the LCs.
Below, we propose a hierarchical contextual online learning algorithm, which is based on algorithms~\cite{ Tekin.Schaar2015, TekinZhangSchaar2014, MuellerAtanSchaarEtAl2016, MuellerAtanSchaarEtAl2016b} for the \textit{contextual multi-armed bandit problem}. 
The algorithm is based on the assumption that a worker's expected performance is similar in similar joint personal and task contexts.
Therefore, by observing the task context, a worker's personal context and her/his performance when requested to complete a task, the worker's context-specific expected performances can be learned and exploited for future worker selection.

\begin{figure}[!t]
\centering
\includegraphics[width=0.36\textwidth]{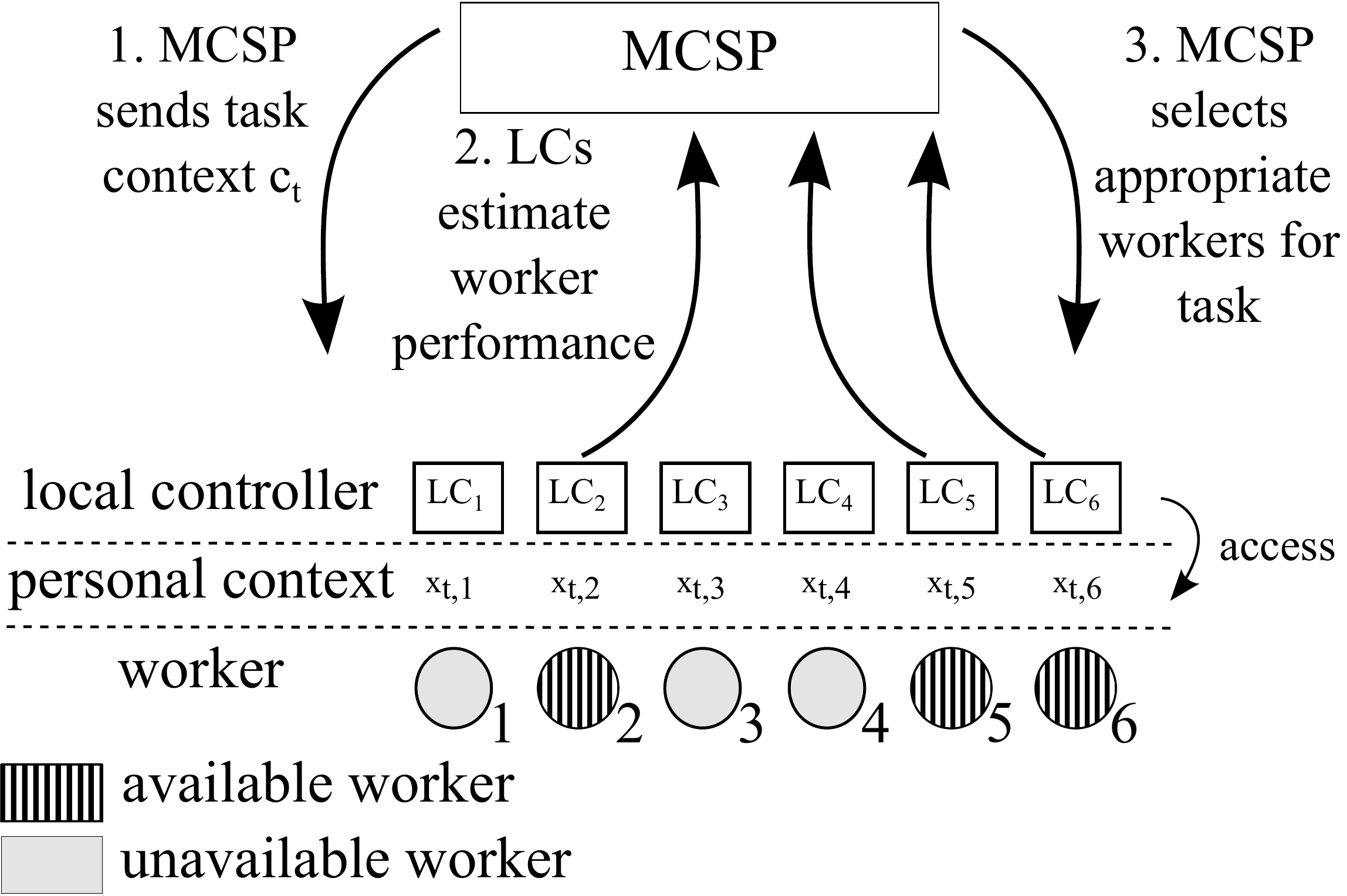}
\caption{Overview of operation of $\PROPOSED$ algorithm for task $t$.}
\label{Fig_hierarchical_learning}
\end{figure}

We call the proposed algorithm \textit{Hierarchical Context-aware Learning} ($\PROPOSED$). Fig.~\ref{Fig_hierarchical_learning} shows an overview of $\PROPOSED$'s operation. In $\PROPOSED$, the MCSP broadcasts the context of each arriving task to the LCs. 
Upon receiving information about a task, an LC first observes its worker's personal context.
If the worker's performance has been observed sufficiently often before given the current joint personal and task context, the LC relies on previous observations to estimate its worker's performance and sends an estimate to the MCSP. 
If its worker's performance has not been observed sufficiently often before, the LC informs the MCSP that its worker has to be explored.
Based on the messages received from the LCs, the MCSP selects a subset of workers.  
The LC of a selected worker requests its worker to complete the task and observes if the worker accepts or declines the task.
If a worker was selected for exploration purposes and accepts the task, the LC additionally observes the quality of the completed task, i.e., depending on the type of quality assessment, the LC gets a quality rating from the task owner or it generates an automatic quality assessment using either local software or the resources of a cloud. 
The reason for only making a quality assessment when a worker was selected for exploration purposes is that quality assessment may be costly.\footnote{If quality assessment is cheap, $\PROPOSED$ can be adapted to always observe worker quality. This may increase the learning speed.} 
Hence, in this way, $\PROPOSED$ keeps the number of costly quality assessments low.

In $\PROPOSED$, a worker's personal contexts, decisions and qualities are only locally stored at the LC.
Thereby, $(i)$~personal context is kept locally, $(ii)$~the required storage space for worker information at the MCSP is kept low, 
$(iii)$~if necessary, task completion and result transmission may be directly handled between the LC and the task owner, $(iv)$~workers receive requests for tasks that are interesting for them and which they are good at, but without the need to share their context information, $(v)$~even though an LC has to keep track of its worker's personal context, decision and quality, the computation and storage overhead for each LC is small.
\begin{algorithm}
\caption{$\PROPOSED$@$\LC$: Local Controller $i$ of Worker $i$.}\label{Algo_LC}
\begin{algorithmic}[1]
\State Receive input from MCSP:  $T$, $\mathcal{C}$, $C$
\State Receive input from worker $i$: $\mathcal{X}_i$, $X_i$
\State Set joint context space $\mathcal{X}_i\times \mathcal{C}$, set $D_i=X_i+C$
\State Set parameter $h_{T,i}\in\mathbb{N}$ and control function $K_i:\{1,...,T\}\rightarrow \mathbb{R}_+$
\State Initialize context partition: Create partition $\mathcal{Q}_{T,i}$ of $[0,1]^{D_i}$ into $(h_{T,i})^{D_i}$ hypercubes of identical size 
\State Initialize counters: For all $q \in \mathcal{Q}_{T,i}$, set
$N_{i,q}=0$ 
\State Initialize estimated performance: For all $q \in \mathcal{Q}_{T,i}$, set $\hat{\theta}_{i,q}=0$
\For{\textbf{each} $t=1,...,T$}
	\If{$i\in\mathcal{W}_t$}
	\State Receive task context $c_t$
	\State Observe worker $i$'s personal context $x_{t,i}$ 
	\State Find the set $q_{t,i} \in \mathcal{Q}_{T,i}$ such that $(x_{t,i},c_t) \in q_{t,i}$
	\If{$N_{i,q_{t,i}}> K_i(t)$}
		\State Send $\mess_i:=\hat{\theta}_{i,q_{t,i}}$ to MCSP
	\Else
		\State Send $\mess_i:=\EXP$ to MCSP
	\EndIf
	\State Wait for MCSP's worker selection 
	\If{MCSP selects worker $i$}
		\State Give task context $c_t$ to worker $i$
		\State Request worker $i$ to complete task $t$
		\State Observe worker $i$'s decision $d$
		\If{$\mess_i==\EXP$}
			\If{$d==1$}
				\State Observe worker $i$'s quality $q$, set $p:=q$
			\Else
				\State Set $p:=0$
			\EndIf
			\State $\hat{\theta}_{i,q_{t,i}} = \frac{\hat{\theta}_{i,q_{t,i}}N_{i,q_{t,i}} + p}							{N_{i,q_{t,i}} + 1}$
			\State	$N_{i,q_{t,i}} = N_{i,q_{t,i}} + 1$
		\EndIf
	\EndIf
	\EndIf
\EndFor
\end{algorithmic}
\end{algorithm}

In more detail, LC~$i$ operates as follows, as given in Alg.~\ref{Algo_LC}.
First, for synchronization purposes, LC~$i$ receives the finite number $T$ of tasks to be considered, the task context space $\mathcal{C}$ and its dimension $C$ from the MCSP.
Moreover, LC~$i$ checks to which of worker~$i$'s context dimensions it has access. 
This defines the personal context space $\mathcal{X}_i$ and its dimension $X_i$.
Then, LC~$i$ sets the joint context space to $\mathcal{X}_i\times \mathcal{C}$ with size $D_i=X_i+C$.
In addition, LC~$i$ has to set a parameter $h_{T,i}\in\mathbb{N}$ and a control function $K_i:\{1,...,T\}\rightarrow \mathbb{R}_+$, which are both described below. 
Next, LC~$i$ initializes a uniform partition $\mathcal{Q}_{T,i}$ of worker $i$'s joint context space $[0,1]^{D_i}$, which consists of $(h_{T,i})^{D_i}$ $D_i$-dimensional hypercubes of equal size $\frac{1}{h_{T,i}}\times \hdots \times\frac{1}{h_{T,i}}$. 
Hence, the parameter $h_{T,i}\in\mathbb{N}$ determines the granularity of the partition of the context space.
Moreover, LC~$i$ initializes a counter $N_{i,q}(t)$ for each hypercube $q\in\mathcal{Q}_{T,i}$.
The counter $N_{i,q}(t)$ represents the number of times before (i.e., up to, but not including) task~$t$, in which worker $i$ was selected to complete a task for exploration purposes when her/his joint context belonged to hypercube~$q$.
Additionally, for each hypercube $q \in \mathcal{Q}_{T,i}$, LC~$i$ initializes the estimate $\hat{\theta}_{i,q}(t)$, which represents the estimated performance of worker $i$ for contexts in hypercube $q$ before task $t$. 

Then, LC~$i$ executes the following steps for each of the tasks $t=1,...,T$.
For an arriving task~$t$, LC~$i$ only takes actions if its worker $i$ is currently available (i.e., $i\in \mathcal{W}_t$). 
If this is the case, LC~$i$ first receives the task context $c_t$ sent by the MCSP.\footnote{A worker being unavailable may mean that she/he is offline. Therefore, we here consider the LC to only take actions if its worker is available.}
Moreover, LC $i$ observes worker $i$'s current personal context $x_{t,i}$ and determines the hypercube from $\mathcal{Q}_{T,i}$ to which the joint context $(x_{t,i},c_t)$ belongs.\footnote{If there are multiple such hypercubes, then, one of them is randomly selected.}
We denote this hypercube by $q_{t,i}\in\mathcal{Q}_{T,i}$. It satisfies  $(x_{t,i},c_t)\in q_{t,i}$.
Then, LC~$i$ checks if worker $i$ has not been selected sufficiently often before when worker $i$'s  joint personal and task context belonged to hypercube $q_{t,i}$.
For this purpose, LC~$i$ compares the counter $N_{i,q_{t,i}}(t)$ with $K_i(t)$, where $K_i:\{1,...,T\}\rightarrow \mathbb{R}_+$ is a deterministic, monotonically increasing control function, set in the beginning of the algorithm. 
On the one hand, if worker~$i$ has been selected sufficiently often before ($N_{i,q_{t,i}}(t)>K_i(t)$), LC~$i$ relies on the estimated performance $\hat{\theta}_{i,q_{t,i}}(t)$, and sends it to the MCSP.
On the other hand, if worker $i$ has not been selected sufficiently often before ($N_{i,q_{t,i}}(t)\leq K_i(t)$), LC~$i$ sends an ``explore'' message to the MCSP. 
The control function $K_i(t)$ is hence needed to distinguish when a worker should be selected for exploration (to achieve reliable estimates) or when the worker's performance estimates are already reliable and can be exploited.
Therefore, the choice of control function is essential to ensure a good result of the learning algorithm, since it determines the trade-off between exploration and exploitation.
Then, LC~$i$ waits for the MCSP to take care of the worker selection.  
If worker $i$ is not selected, LC~$i$ does not take further actions.
However, if the MCSP selects worker $i$, LC $i$ gives the task context information $c_t$ to worker $i$ via the application's user interface and requests worker $i$ to complete the task. Then, 
LC $i$ observes whether worker $i$ declines or accepts the task.
If worker $i$ was selected for exploration purposes, LC~$i$ makes an additional counter update.
For this, if worker $i$ accepted the task, LC~$i$ additionally observes worker $i$'s quality in completing the task (e.g., by receiving a quality rating from the task owner or by generating an automatic quality assessment) and sets the observed performance to the observed quality.
If worker $i$ declined the task, LC~$i$ sets the observed performance to $0$.
Then, based on the observed performance, LC~$i$ computes the estimated performance $\hat{\theta}_{i,q_{t,i}}(t+1)$ for hypercube $q_{t,i}$ and the counter $N_{i,q_{t,i}}(t+1)$.
Note that in Alg.~\ref{Algo_LC}, the argument~$t$ is omitted from counters $N_{i,q}(t)$ and estimates $\hat{\theta}_{i,q}(t)$ since it is not necessary to store their respective previous values.

By definition of $\PROPOSED$, the estimated performance $\hat{\theta}_{i,q}(t)$ corresponds to the product of $(i)$~the relative frequency with which worker $i$ accepted tasks when the joint context belonged to hypercube $q$ and $(ii)$~the average quality in completing these tasks. 
Formally, $\hat{\theta}_{i,q}(t)$  is computed as follows. 
Let $\mathcal{E}_{i,q}(t)$ be the set of observed performances of worker $i$ before task $t$ when worker $i$ was selected for a task and the joint context was in hypercube $q$. 
If before task $t$, worker $i$'s performance has never been observed before for a joint context in hypercube $q$, we have $\mathcal{E}_{i,q}(t)= \emptyset$ and $\hat{\theta}_{i,q}(t):=0$.
Otherwise, the estimated performance is given by 
$\hat{\theta}_{i,q}(t):=\frac{1}{|\mathcal{E}_{i,q}(t)|} \sum_{p\in \mathcal{E}_{i,q}(t)} p$.
However, in $\PROPOSED$, the set $\mathcal{E}_{i,q}(t)$ does not appear, since the estimated performance $\hat{\theta}_{i,q}(t)$ can be computed based on $\hat{\theta}_{i,q}(t-1)$, $N_{i,q}(t-1)$ and on the performance for task $t-1$.

\begin{algorithm}
\caption{$\PROPOSED$@$\MCSP$: Worker Selection at MCSP.}\label{Algo_MCSP}
\begin{algorithmic}[1]
\State Send input to LCs:  $T$, $\mathcal{C}$, $C$
\For{\textbf{each} $t=1,...,T$}
	\State Receive task $t=(b_t,c_t)$
	\State Compute $m_t=\lfloor \frac{b_t}{e_t} \rfloor$
	\State Set $\mathcal{W}_{t}=\emptyset$
	\State Set $\mathcal{W}_{t}^{\ue}=\emptyset$
	\State Broadcast task context $c_t$
	\For{\textbf{each} $i=1,...,W$}
		\If{Receive $\mess_i$ from LC $i$}
			\State $\mathcal{W}_{t}=\mathcal{W}_{t}\cup\{i\}$
			\If{$\mess_i==\EXP$}
				\State $\mathcal{W}_{t}^{\ue}=\mathcal{W}_{t}^{\ue}\cup \{i\}$
			\EndIf
		\EndIf
	\EndFor
	\State Compute $W_t=|\mathcal{W}_{t}|$ 
	\If{$W_t\leq m_t$}\Comment{SELECT ALL}
		\State Select all $W_t$ workers from $\mathcal{W}_{t}$
		\Else
			\State Compute $n_{\ue,t}=|\mathcal{W}_{t}^{\ue}|$ 
			\If{$n_{\ue,t}==0$}\Comment{EXPLOITATION}
				\State  \parbox[t]{\dimexpr\linewidth-4.5em}{Rank workers in $\mathcal{W}_t$ according to estimates from $(\mess_i)_{i\in\mathcal{W}_t}$\strut}
				\State Select the $m_t$ highest ranked workers 
			\Else\Comment{EXPLORATION}
				\If{$n_{\ue,t}\geq m_t$ }
					\State Select $m_t$ workers randomly from $\mathcal{W}_{t}^{\ue}$
				\Else
					\State Select the $n_{\ue,t}$ workers from $\mathcal{W}_{t}^{\ue}$
					\State \parbox[t]{\dimexpr\linewidth-6em}{Rank workers in $\mathcal{W}_t\setminus \mathcal{W}_{t}^{\ue}$ according to estimates from $(\mess_i)_{i\in\mathcal{W}_t\setminus \mathcal{W}_{t}^{\ue}}$\strut}
				\State \parbox[t]{\dimexpr\linewidth-6em}{Select the $(m_t-n_{\ue,t})$ highest ranked workers\strut}
				\EndIf
			\EndIf
	\EndIf
	\State Inform LCs of selected workers
\EndFor
\end{algorithmic}
\end{algorithm}

In $\PROPOSED$, the MCSP is responsible for the worker selection, which it executes according to Alg.~\ref{Algo_MCSP}.
First, for synchronization purposes, the MCSP sends the finite number~$T$ of tasks to be considered, the task context space~$\mathcal{C}$ and its dimension~$C$ to the LCs.
Then, for each arriving task $t=(b_t,c_t)$, the MCSP computes the maximum number  $m_t$ of workers, based on the budget $b_t$ and the price $e_t$ per worker. In addition, the MCSP initializes two sets. 
The set $\mathcal{W}_t$ represents the set of available workers when task $t$ arrives,
while $\mathcal{W}_{t}^{\ue}$ is the so-called \textit{set of under-explored workers}, which contains all available workers that have not been selected sufficiently often before.
After broadcasting the task context $c_t$, the MCSP waits for messages from the LCs. 
If the MCSP receives a message from an LC, it adds the corresponding worker to the set $\mathcal{W}_t$ of available workers.
Moreover, in this case the MCSP additionally checks if the received message is an ``explore'' message.
If this is the case, the MCSP adds the corresponding worker to the set $\mathcal{W}_{t}^{\ue}$ of under-explored workers.
Note that according to~Alg.~\ref{Algo_LC} and Alg.~\ref{Algo_MCSP}, the set of under-explored workers is hence given by
\begin{align}\label{Set_underexplored}
\mathcal{W}_{t}^{\ue}&=\{i\in \mathcal{W}_t: N_{i,q_{t,i}}(t)\leq K_i(t)\}.
\end{align}
Next, the MCSP calculates the number $W_t$ of available workers. 
If $W_t\leq m_t$, i.e., at most the required number of workers are available, the MCSP enters a \textit{select-all-workers phase} and selects all available workers to complete the task.
Otherwise, the MCSP continues by calculating the number $n_{\ue,t}:=|\mathcal{W}_{t}^{\ue}|$ of under-explored workers.
If there is no under-explored worker, the MCSP enters an \textit{exploitation phase}.
It ranks the available workers in $\mathcal{W}_t$ according to the estimated performances, which it received from their respective LCs. 
Then, the MCSP selects the $m_t$ highest ranked workers.
By this procedure, the MCSP is able to use context-specific estimated performances without actually observing the workers' personal contexts.
If there are under-explored workers, the MCSP enters an \textit{exploration phase}.
These phases are needed, such that all LCs are able to update their estimated performances sufficiently often.
Here, two different cases may occur, depending on the number $n_{\ue,t}$ of under-explored workers.
Either the number $n_{\ue,t}$ of under-explored workers is at least $m_t$, in which case the MCSP selects $m_t$ under-explored workers at random.
Or the number $n_{\ue,t}$ of under-explored workers is smaller than $m_t$, in which case the MCSP selects all $n_{\ue,t}$ under-explored workers. 
Since it should select $m_t-n_{\ue,t}$ additional workers, it ranks the available sufficiently-explored workers according to the estimated performances, which it received from their respective LCs. 
Then, the MCSP additionally selects the $(m_t-n_{\ue,t})$ highest ranked workers.
In this way, additional exploitation is carried out in exploration phases, when the number of under-explored workers is small.
After worker selection, the MCSP informs the LCs of selected workers that their workers should be requested to complete the task. 
Note that since the MCSP does not have to keep track of the workers' decisions, the LCs may handle the contact to the task owner directly (e.g., the task owner may send detailed task instructions directly to the LC; after task completion, the LC may send the result to the task owner).

\section{Theoretical Analysis}\label{Sec_analysis}
\subsection{Upper Bound on Regret}

The performance of $\PROPOSED$ is evaluated by analyzing its regret, see Eq.~\eqref{Eq_regret}, with respect to the centralized oracle. 
In this section, we derive a sublinear bound on the regret, i.e., we show that $R(T) = O(T^{\gamma})$ with some $\gamma<1$ holds. 
Hence, our algorithm converges to the centralized oracle for $T\rightarrow \infty$, since $\lim_{T\rightarrow \infty}\frac{R(T)}{T} = 0$ holds.
The regret bound is derived based on the assumption that under a similar joint personal and task context, a worker's expected performance is also similar.
This assumption can be formalized as follows.\footnote{Note that our algorithm can also be applied to data, which does not satisfy this assumption. In this case, the regret bound may, however, not hold.}
\begin{Assumption}[H\"older continuity assumption] \label{Ass_Hoelder}
There exists $L>0$, $0<\alpha\leq 1$ such that for all workers $i\in\mathcal{W}$ and for all joint contexts $(x,c), (\tilde{x},\tilde{c}) \in \mathcal{X}_i\times \mathcal{C} \equiv [0,1]^{D_i}$, it holds that
\begin{align*}
\left|\theta_i(x,c)-\theta_i(\tilde{x},\tilde{c})\right|\leq L ||(x,c)-(\tilde{x},\tilde{c})||_{i}^{\alpha},
\end{align*}
where $||\cdot||_{i}$ denotes the Euclidean norm in $\mathbb{R}^{D_i}$. 
\end{Assumption}
The theorem given below shows that the regret of $\PROPOSED$ is sublinear in the time horizon~$T$.
\begin{Theorem}[Bound for $R(T)$]\label{Theorem}
Given that Assumption~\ref{Ass_Hoelder} holds, when LC $i$, $i \in {\cal W}$, runs Alg.~\ref{Algo_LC} with parameters
$K_i(t) = t^{\frac{2\alpha}{3 \alpha + D_i}} \log (t)$, $t=1,...,T$, and $h_{T,i} = \ceil{T^{\frac{1}{3\alpha + D_i}}}$, and the MCSP runs Alg.~\ref{Algo_MCSP}, the regret $R(T)$ is bounded by
\begin{align*}
R(T) &\leq  q_{\max}W \sum_{i\in\mathcal{W}} 2^{D_i} \left(\log(T) T^{\frac{2\alpha + D_i}{3 \alpha + D_i}} + T^{\frac{D_i}{3 \alpha + D_i}}\right) \\
 & \quad + \sum_{i\in\mathcal{W}} \frac{2q_{\max}}{(2\alpha + D_i)/(3 \alpha + D_i)}T^{\frac{2\alpha + D_i}{3 \alpha + D_i}}\\
 &\quad + q_{\max} W^2 \frac{\pi^2}{3} + 2 \sum_{i\in\mathcal{W}} L D_i^{\frac{\alpha}{2}}T^{\frac{2\alpha + D_i}{3 \alpha + D_i}}.
 \end{align*}
Hence, the leading order of the regret is $O\left(q_{\max}W^2 T^{\frac{2\alpha + D_{\max}}{3\alpha + D_{\max}}} \log(T) \right)$, where $D_{\max}:=\max_{i\in\mathcal{W}}D_i$.
\end{Theorem}
The proof of Theorem \ref{Theorem} is given in Appendix~\ref{App_regret}. 
Theorem~\ref{Theorem} shows that $\PROPOSED$ converges to the centralized oracle in the sense that when the number $T$ of tasks goes to infinity, the averaged regret $\frac{R(T)}{T}$ diminishes.
Moreover, since Theorem~\ref{Theorem} is applicable for any finite number~$T$ of tasks, it characterizes $\PROPOSED$'s speed of learning.

\subsection{Local Storage Requirements}

The required local storage size in the mobile device of a worker is determined by the storage size needed when the LC executes Alg.~\ref{Algo_LC}. 
In Alg.~\ref{Algo_LC}, LC $i$ stores the counters $N_{i,q}$ and estimates $\hat{\theta}_{i,q}$ for each $q\in\mathcal{Q}_{T,i}$. 
Using the parameters from Theorem~\ref{Theorem}, the number of hypercubes in the partition $\mathcal{Q}_{T,i}$ is $(h_{T,i})^{D_i} = \ceil{T^{\frac{1}{3\alpha + D_i}}}^{D_i}\leq (1+T^{\frac{1}{3\alpha + D_i}})^{D_i}$.
Hence, the number of variables to store in the mobile device of worker~$i$ is upper bounded by $2 \cdot (1+T^{\frac{1}{3\alpha + D_i}})^{D_i}$.
Hence, the required storage depends on the number $D_i= X_i + C$ of context dimensions.
If the worker allows access to a high number $X_i$ of personal context dimensions and/or the number $C$ of task context dimensions is large, the algorithm learns the worker's context-specific performance with finer granularity and therefore the assigned tasks are more personalized, but also the required storage size will increase.

\subsection{Communication Requirements}

The communication requirements of $\PROPOSED$ can be deduced from its main operation steps.
For each task $t$, the MCSP broadcasts the task context to the LCs, which is one vector of dimension $C$ (i.e., $C$ scalars), assuming that the broadcast reaches all workers in a single transmission.
Then, the LCs of available workers send their workers' estimated performances to the MCSP. 
This corresponds to $W_t$ scalars to be transmitted (one scalar sent by each LC of an available worker). 
Finally, the MCSP informs selected workers about its decision, which corresponds to $m_t$ scalars sent by the MCSP. 
Hence, for task~$t$, in sum, $C + W_t + m_t$ scalars are transmitted. Among these, $C + m_t$ scalars are transmitted by the MCSP and one scalar is transmitted by each mobile device of an available worker.

We now compare the communication requirements of $\PROPOSED$ and of its centralized version, called here CCL. In CCL, for each task, the personal contexts of available workers are gathered in the MCSP, which then selects workers based on the task and personal contexts and informs selected workers about its decision. 
The communication requirements of CCL are as follows:
For each task $t$, the LC of each available worker~$i$ sends the current worker context to the MCSP, which is a vector of dimension $D_i$ (i.e., $D_i$ scalars). 
Hence, in sum, $\sum_{i\in\mathcal{W}_t}D_i$ scalars are transmitted.
After worker selection, the MCSP requests selected workers to complete the task, which corresponds to $m_t$ scalars sent by the MCSP. 
Moreover, the MCSP broadcasts the task context to the selected workers, which is one vector of dimension $C$ (i.e., $C$ scalars), assuming that the broadcast reaches all addressed workers in a single transmission.
Hence, in total, $\sum_{i\in\mathcal{W}_t}D_i + m_t  + C$ scalars are transmitted for task $t$. Among these, $C+m_t$ scalars are transmitted by the MCSP and $D_i$ scalars are transmitted by each mobile device of an available worker.

We now compare $\PROPOSED$ with CCL. 
The mobile device of any worker $i\in\mathcal{W}$ with $D_i>1$ has to transmit less using $\PROPOSED$ than using CCL. Moreover, under the assumption that any broadcast reaches all addressed workers using one single transmission, if $D_i\geq 1$ for all $i\in\mathcal{W}$ (i.e., each worker gives access to at least one personal context), the sum communication requirements (for all mobile devices and the MCSP in sum) of $\PROPOSED$ are at most as high as that of CCL.

\subsection{Worker Quality Assessment Requirements}

Observing a worker's quality might be costly. 
$\PROPOSED$ explicitly takes this into account by only requesting a quality assessment if a worker is selected for exploration purposes. 
Here, we give an upper bound on the number $A_i(T)$ of quality assessments per worker up to task $T$.
\begin{Corollary}[Bound for number of quality assessments up to task~$T$]\label{Corollary_quality_assessments}
Given that Assumption~\ref{Ass_Hoelder} holds, when LC $i$, $i \in {\cal W}$, runs Alg.~\ref{Algo_LC} with the parameters given in Theorem~\ref{Theorem}, and the MCSP runs Alg.~\ref{Algo_MCSP}, the number $A_i(T)$ of quality assessments of each worker $i$ up to task $T$ is upper bounded by
\begin{align*}
A_i(T)\leq (1+T^{\frac{1}{3 \alpha + D_i}})^{D_i} \left(1 + \log(T) T^{\frac{2\alpha}{3 \alpha + D_i}} \right).
\end{align*}
\end{Corollary}
The proof of Corollary~\ref{Corollary_quality_assessments} is given in Appendix~\ref{App_thm_qual_ass}. 
From Corollary~\ref{Corollary_quality_assessments}, we see that the number of quality assessments per worker is sublinear in $T$. 
Hence, it holds $\lim_{T\rightarrow \infty}\frac{A_i(T)}{T}=0$, so that for $T\rightarrow \infty$, the average rate of quality assessments approaches zero.

\section{Numerical Results}\label{Sec_num_results}
We evaluate $\PROPOSED$ by comparing its performance with various algorithms based on synthetic and real data.

\subsection{Reference Algorithms}
The following algorithms are used for comparison.
\begin{itemize}
\item The \textit{(Centralized) Oracle} has perfect a priori knowledge about context-specific expected performances and knows the current contexts of available workers. 
\item \textit{LinUCB} assumes that the expected performance of a worker is linear in its context~\cite{LiChuLangfordEtAl2010},~\cite{ChuLiReyzinEtAl2011}.
Based on a linear reward function over contexts and previously observed context-specific worker performances, for each task, LinUCB chooses the $m_t$ available workers with highest estimated upper confidence bounds on their expected performance. 
LinUCB has an input parameter $\lambda_{\LinUCB}$, controlling the influence of the confidence bound. LinUCB is used in \cite{HassanCurry2014} for task assignment in spatial CS. 
\item \textit{AUER} \cite{KleinbergNiculescu-MizilSharma2010} is an extension of the well-known UCB algorithm \cite{AuerCesa-BianchiFischer2002} to the sleeping arm case.
It learns from previous observations of worker performances, but without taking into account context information. Based on the history of previous observations of worker performances, AUER selects the $m_t$ available workers with highest estimated upper confidence bounds on their expected performance.  
AUER has an input parameter $\lambda_{\AUER}$, which controls the influence of the confidence bound.
\item \textit{$\epsilon$-Greedy} selects a random subset of available workers with a probability of $\epsilon\in(0,1)$. With a probability of $(1-\epsilon)$, $\epsilon$-Greedy selects the $m_t$ available workers with highest estimated performance. The estimated performance of a worker is computed based on the history of previous performances~\cite{AuerCesa-BianchiFischer2002}, but without taking into account context. 
\item \textit{Myopic} only learns from the last interaction with each worker. For task $1$, it selects a random subset of $m_1$ workers. For each of the following tasks, it checks which of the available workers have previously accepted a task. If more than $m_t$ of the available workers have accepted a task when requested the last time, Myopic selects out of these workers the $m_t$ workers with the highest performance in their last completed task. Otherwise, Myopic selects all of these workers and an additional subset of random workers so that in total $m_t$ workers are selected.
\item \textit{Random} selects a random subset of $m_t$ available workers for each task $t$.
\end{itemize}
Note that, if an algorithm originally would have selected only one worker per task, we adapted it to select $m_t$ workers per task.
Also, above, we described the behavior of the algorithms for the case $m_t< W_t$. In the case of $m_t\geq W_t$, we adapted each algorithm such that it selects all available workers.
Moreover, while we used standard centralized implementations of the reference algorithms, they could also be decoupled to a hierarchical setting like $\PROPOSED$.

\subsection{Evaluation Metrics}
Each algorithm is run over a sequence of tasks $t=1,...,T$ and its result is evaluated using the following metrics.
We compute the \textit{cumulative worker performance at $T$} achieved by an algorithm, which is the cumulative sum of performances by all selected workers up to (and including) task $T$.
Formally, if the set of selected workers of an algorithm~$A$ for task $t$ is $\{s^A_{t,j}\}_{j=1,...,\min\{m_{t},W_{t}\}}$ and $p_{s^A_{t,j}}(t)$ is the observed performance of worker $s^A_{t,j}$, the cumulative worker performance at $T$ achieved by algorithm~$A$ is
\begin{align*}
\Gamma_T(A):=\sum_{t=1}^T\sum_{j=1}^{\min\{m_{t},W_{t}\}} p_{s^A_{ t,j}}(t).
\end{align*}
As a function of the arriving tasks, we compute the \textit{average worker performance up to~$t$} achieved by an algorithm, which is the average performance of all selected workers up to task~$t$.
Formally, it is defined by
\begin{align*}
\frac{1}{\sum_{\tilde{t}=1}^t \min\{m_{\tilde{t}},W_{\tilde{t}}\}}\sum_{\tilde{t}=1}^t\sum_{j=1}^{\min\{m_{\tilde{t}},W_{\tilde{t}}\}} p_{s^A_{\tilde{t},j}}(\tilde{t}).
\end{align*}

\subsection{Simulation Setup}\label{Subsec_sim_setup}
We evaluate the algorithms using synthetic and real data. 
The difference between the two approaches lies in the arrival process of workers and their contexts. 
To produce synthetic data, we generate workers and their contexts based on some predefined distributions as described below.
In case of real data, similar to, e.g.,~\cite{KazemiShahabi2012, HassanCurry2014, ZhengChen2017}, we use a data set from Gowalla~\cite{ChoMyersLeskovec2011}. 
Gowalla is a location-based social network where users share their location by checking in at ``spots'', i.e., certain places in their vicinity.
We use the check-ins to simulate the arrival process of workers and their contexts.
The Gowalla data set consists of 6,442,892 check-ins of 107,092 distinct users over the period of February 2009 to October 2010.
Each entry of the data set consists of the form (User ID, Check-in Time, Latitude, Longitude, Location ID).
Similar to~\cite{ZhengChen2017}, we first extract the check-ins in New York City, which leaves a subset of 138,954 check-ins of 7,115 distinct users at 21,509 distinct locations.
This resulting Gowalla-NY data set is used below.
\begin{figure}[!t]
\centering
\subfigure[Check-ins.]{\label{Fig_check-ins}\includegraphics[width=0.23\textwidth]{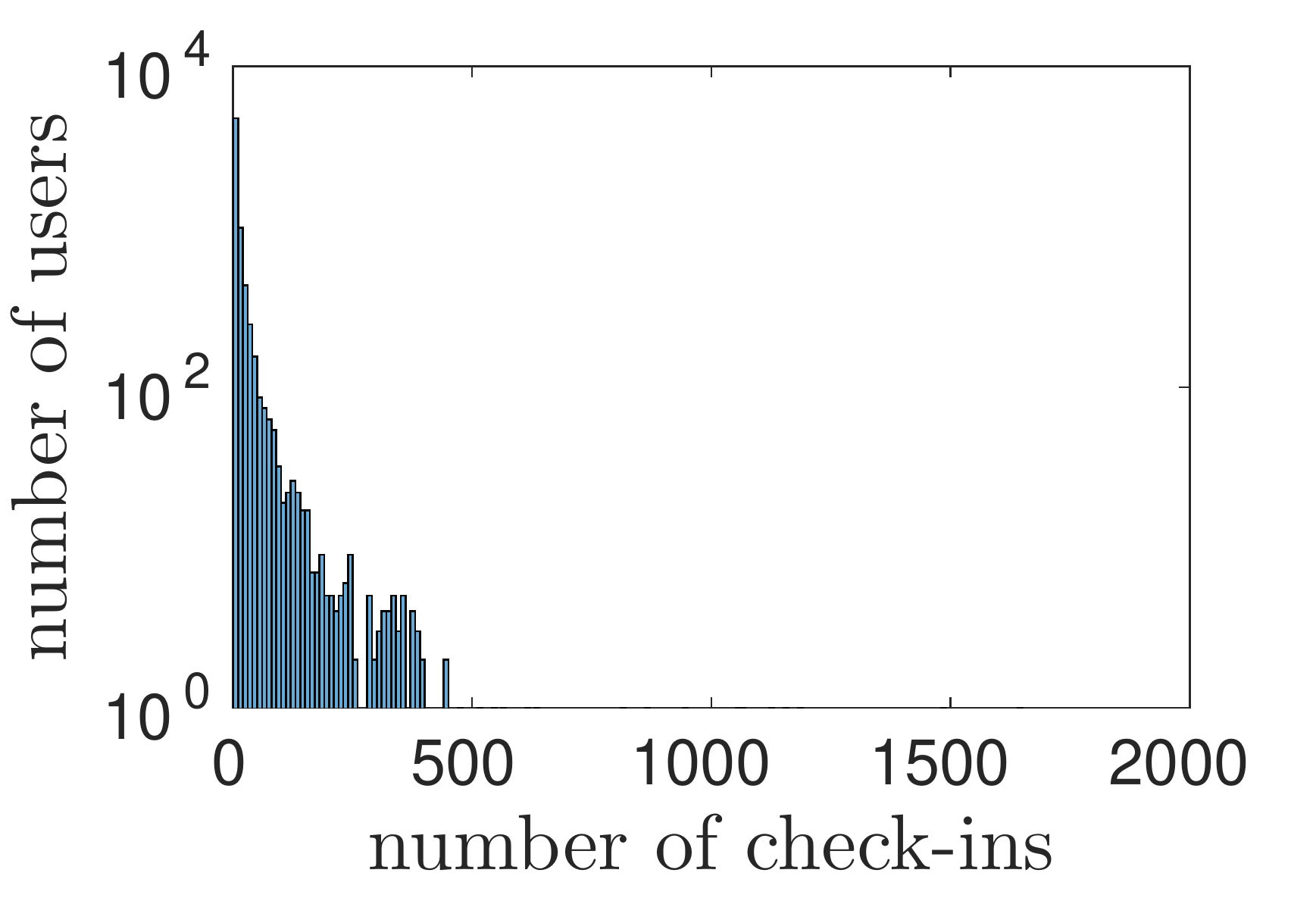}}
\hfil
\subfigure[Visited locations.]{\label{Fig_locations}\includegraphics[width=0.23\textwidth]{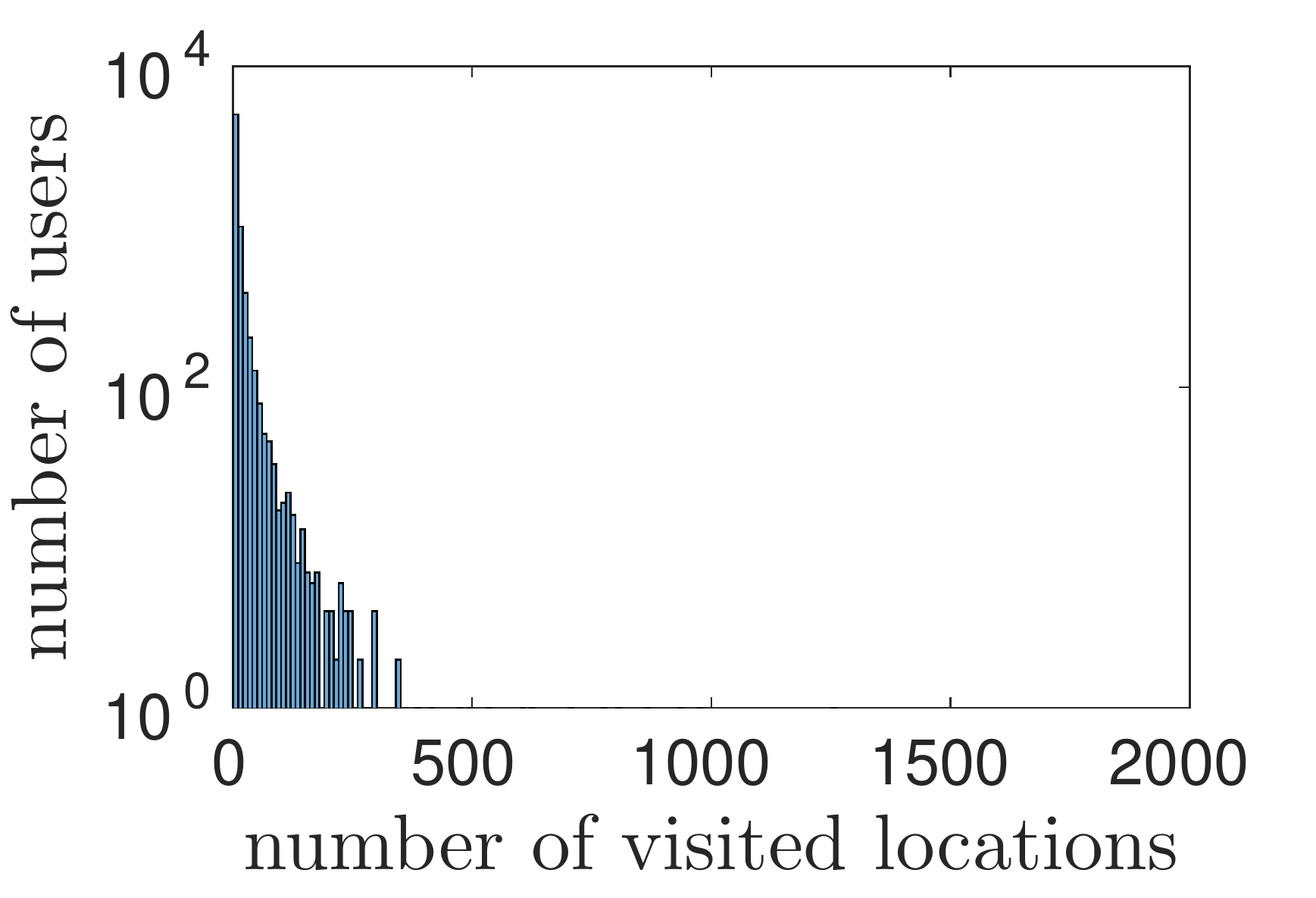}}
\caption{Statistics of used Gowalla-NY data set.}
\label{Fig_statistics_gowalla}
\end{figure}
Fig.~\ref{Fig_check-ins} and Fig.~\ref{Fig_locations} show the distributions of the number of check-ins and the number of distinct locations visited by the users in the Gowalla-NY data set, respectively. 

For both synthetic and real data, we simulate an MCSP, to which a set of $W=100$ workers belongs. 
For synthetic data, $100$ workers are created in the beginning. 
For real data, we randomly select $100$ users from the Gowalla-NY data set, which represent the $100$ workers of the MCSP. Then we use this reduced Gowalla-NY data set containing the check-ins of $100$ users.
On top of this, the simulation is modeled as follows: 
\subsubsection{Task Properties}
The task context is assumed to be uniformly distributed in $\mathcal{C}=[0,1]$ (i.e., $C=1$).
Task owners have to pay a fixed price of $e=0.75$ or $e=1$ per requested worker that completes the task when the task context lies in $[0,0.5]$ or $(0.5,1]$, respectively.
The quality of a completed task lies in the range $q_{\min}=0$ and $q_{\max}=5$. 
The task budget is sampled from a normal distribution with expected value $20$ and standard deviation of $5$, truncated between $1$ and $100$. 

\subsubsection{Worker Availability}
For synthetic data, we let each worker be available with a probability of $ \rho = 0.7$ (default value) for each arriving task. 
For the real data, we use a Binomial distribution with parameters $W=100$ and $\rho=0.7$ (default value) to sample the number of available workers $W_t$ for an arriving task.\footnote{In this way, the number of available workers in our experiments using the real and the synthetic data are distributed in the same way.}
Having sampled $W_t$, we randomly draw samples from the reduced Gowalla-NY data set (consisting of the check-ins of $100$ users) until these samples contain $W_t$ distinct users.
These $W_t$ sampled users correspond to the available workers (i.e., users with higher number of check-ins in the reduced Gowalla-NY data set translate to workers that are more often available for the MCSP).

\subsubsection{Worker Context}
The personal context space of an available worker $i$ is set to $\mathcal{X}_i=[0,1]^2$ (i.e., $X_i=2$).
The first personal context dimension refers to the worker's battery state, which is sampled from a uniform distribution in $[0,1]$. 
The second personal context dimension refers to the worker's location, which is sampled differently in case of synthetic and real data. For synthetic data, the worker's location is sampled from $5$ different (personal) locations, using a weighted discrete distribution with probabilities $\{\frac{1}{2},\frac{1}{3},\frac{1}{12},\frac{1}{24},\frac{1}{24}\}$ to represent the fact that workers may use the MCS application different amounts of time in different places (e.g., at home more often than at work). 
For real data, we set the worker's location to be the check-in location of the respective user from the sample.\footnote{If a user was sampled several times until we sampled $W_t$ distinct users, we choose her/his first sampled check-in location.}

\subsubsection{Expected Worker Performance}
We use two different models to generate expected worker performance.
\paragraph{Discrete Performance Model}
The joint personal and task context space $\mathcal{X}_i\times \mathcal{C}$ (of dimension $D_i = 3$) is split into a uniform grid. For synthetic data, the space is split into $5$ identical parts along each of the $3$ dimensions, i.e., $5\cdot 5\cdot 5=125$ subsets are created. 
For real data, along the dimensions of task context and battery state, the context space is split into $5$ identical parts each, but along the dimension of location context, the context space is split into $l_i$ identical parts, where $l_i$ corresponds to the number of distinct locations visited by the corresponding user from the reduced Gowalla-NY data set. Hence, $5\cdot 5 \cdot l_i$ subsets are created. 
Then, for both synthetic and real data, in each of the subsets, the expected performance of a worker is a priori sampled uniformly at random from $[0,5]$. 
Note that for the real data, since the expected performance differs per visited location, workers with higher number of visited locations have a higher number of different context-specific performances.

\paragraph{Hybrid Performance Model} 
We assume a continuous dependency of the expected performance on two of the context dimensions. Let $x_{i}^{(1)}$ and $x_{i}^{(2)}$ be worker $i$'s battery state and location, respectively, and let $c$ be the task context. We assume that the expected performance $\theta_i$ of worker $i$ is given by
\begin{align*}
\theta_i\left(c, x_{i}^{(1)}, x_{i}^{(2)}\right) = q_{\max}\cdot w_i\left(x_{i}^{(2)}\right) \cdot  \bar{f}_{\mu_i, \sigma_i^2}\left(c\right) \cdot \sqrt{x_{i}^{(1)}} ,
\end{align*}
where $w_i\left(x_{i}^{(2)}\right)$ is a (discrete) location-specific weighting factor that is a priori sampled uniformly between $[0.5,1]$ for each of worker $i$'s (finitely many) locations. Moreover, $\bar{f}_{\mu_i, \sigma_i^2}$ is a Gaussian probability density function with mean $\mu_i$ and standard deviation $\sigma_i$, which is normalized such that its maximum value equals 1. For worker $i$, the mean $\mu_i$ is a priori sampled uniformly from $[0.1,0.9]$ and the standard deviation is set to $\sigma_i = 0.1\cdot \mu_i$. Hence, the expected performance is a continuous function of task context and battery state. The hybrid model has the following intuition: The expected performance of a worker is location-specific. Along the task context, the expected performance varies according to a worker-specific Gaussian distribution, i.e., each worker performs well at a specific type of tasks. Finally, the expected performance grows monotonically with the battery state, i.e., with more battery available, workers are more likely to perform well at tasks.

\subsubsection{Instantaneous Worker Performance}
For each occurring joint worker and task context, the instantaneous performance of a worker is sampled by adding noise uniformly sampled from $[-1,1]$ to the expected performance in the given context (the noise interval is truncated to a smaller interval if the expected performance lies close to either $0$ or $q_{\max}$).

\subsection{Parameter Selection}
$\PROPOSED$, LinUCB, AUER and $\epsilon$-Greedy require input parameters. In order to find appropriate parameters, we generate $20$ synthetic instances using the discrete performance model. Each instance consists of a sequence of $T=10,000$ task and worker arrivals sampled according to Sec.~\ref{Subsec_sim_setup}.
Then, we run each algorithm with different parameters on these instances.
Note that for $\PROPOSED$, we set $\alpha=1$, choose $h_{T,i} = \ceil{T^{\frac{1}{3 + D_i}}}$, $i\in\mathcal{W}$, as in Theorem~\ref{Theorem}, and set the control function to $K_i(t) = f t^{\frac{2\alpha}{3 \alpha + D_i}} \log (t)$, $t=1,...,T$, where the factor $f\in (0,1]$ is included to reduce the number of exploration phases. Then, we search for an appropriate $f$.
Table~\ref{Table_parameters} shows the parameters at which each of the algorithms on average performed best, respectively. 
These parameters are used in all of the following simulations.
\begin{table}[!t]
\renewcommand{\arraystretch}{1.1}
\caption{Choice of Parameters for Different Algorithms.}
\label{Table_parameters}
   \centering
      \begin{tabular}{|c|c|c|}     
      \hline
      Algorithm & Parameter & Selected Value\\
      \hline
		$\PROPOSED$ & $f$ & $0.003$\\
      \hline
		LinUCB & $\lambda_{\LinUCB}$ & $1.5$\\
      \hline
		AUER & $\lambda_{\AUER}$ & $0.5$\\
      \hline
      $\epsilon$-Greedy & $\epsilon$ & $0.01$\\
      \hline
      \end{tabular}
\end{table}

\subsection{Results under the Discrete Performance Model}\label{Subsec_DiscreteModel}

First, we generate $100$ synthetic and $100$ real instances, in both cases using $\rho=0.7$ and the discrete performance model. Each instance consists of a sequence of $T=10,000$ task and worker arrivals sampled according to Sec.~\ref{Subsec_sim_setup}.
Then, we run the algorithms on these instances and average the results.

For both synthetic and real data, Table~\ref{Table_cum_perf} compares the cumulative worker performance at~$T$ of an algorithm $A$ with the one of $\PROPOSED$, by displaying $\Gamma_T(A)/\Gamma_T(\PROPOSED)$.
\begin{table}[!t]
\renewcommand{\arraystretch}{1.1}
\caption{Comparison of cumulative worker performance at $T$ for $\rho=0.7$ under the discrete performance model. For an algorithm  $A$, the table shows  $\Gamma_T(A)/\Gamma_T(\PROPOSED)$.}
\label{Table_cum_perf}
   \centering
      \begin{tabular}{|c|c|c|}     
      \hline
      Algorithm & Synthetic Data & Real Data\\
      \hline
		Oracle & $1.04$ & $1.20$\\
      \hline
		$\PROPOSED$ & $1.00$ & $1.00$\\
      \hline
		LinUCB & $0.69$ & $0.78$\\
      \hline
		AUER & $0.68$ & $0.77$\\
      \hline
      $\epsilon$-Greedy & $0.68$ & $0.76$\\
\hline
      Myopic & $0.64$ & $0.74$\\
      \hline
            Random & $0.64$ & $0.73$\\
      \hline
      \end{tabular}
\end{table}
As expected, while Oracle outperforms all other algorithms due to its a priori knowledge, Random gives a lower bound on the achievable cumulative performance. 
$\PROPOSED$ clearly outperforms LinUCB, AUER, $\epsilon$-Greedy and Myopic, even though $\PROPOSED$ observes worker performance only when requesting a worker for exploration purposes, while the other algorithms have access to worker performance whenever a worker is requested. This is due to the fact that $\PROPOSED$ smartly exploits context. 
Moreover, $\PROPOSED$ reaches a result close to the Oracle. 
In contrast, LinUCB, AUER, $\epsilon$-Greedy and Myopic perform by far worse and lie close to the result of Random. 
This shows that algorithms which either do not take context into account (i.e., AUER, $\epsilon$-Greedy and Myopic) or have a linearity assumption between context and performance (i.e., LinUCB), cannot cope with the non-linear dependency of expected worker performance on context.
Comparing synthetic and real data, $\PROPOSED$ has a better performance on the synthetic data, but it still reaches a good result on the real data, even though using real data, each worker has her/his own diversity in context arrival and hence in expected performance (since users in the Gowalla-NY data set have different numbers of visited check-in locations).

\begin{figure}[!t]
\centering
\subfigure[Experiments with synthetic data.]{\label{Fig_results_syn_av_perf_prav70}\includegraphics[width=0.46\textwidth]{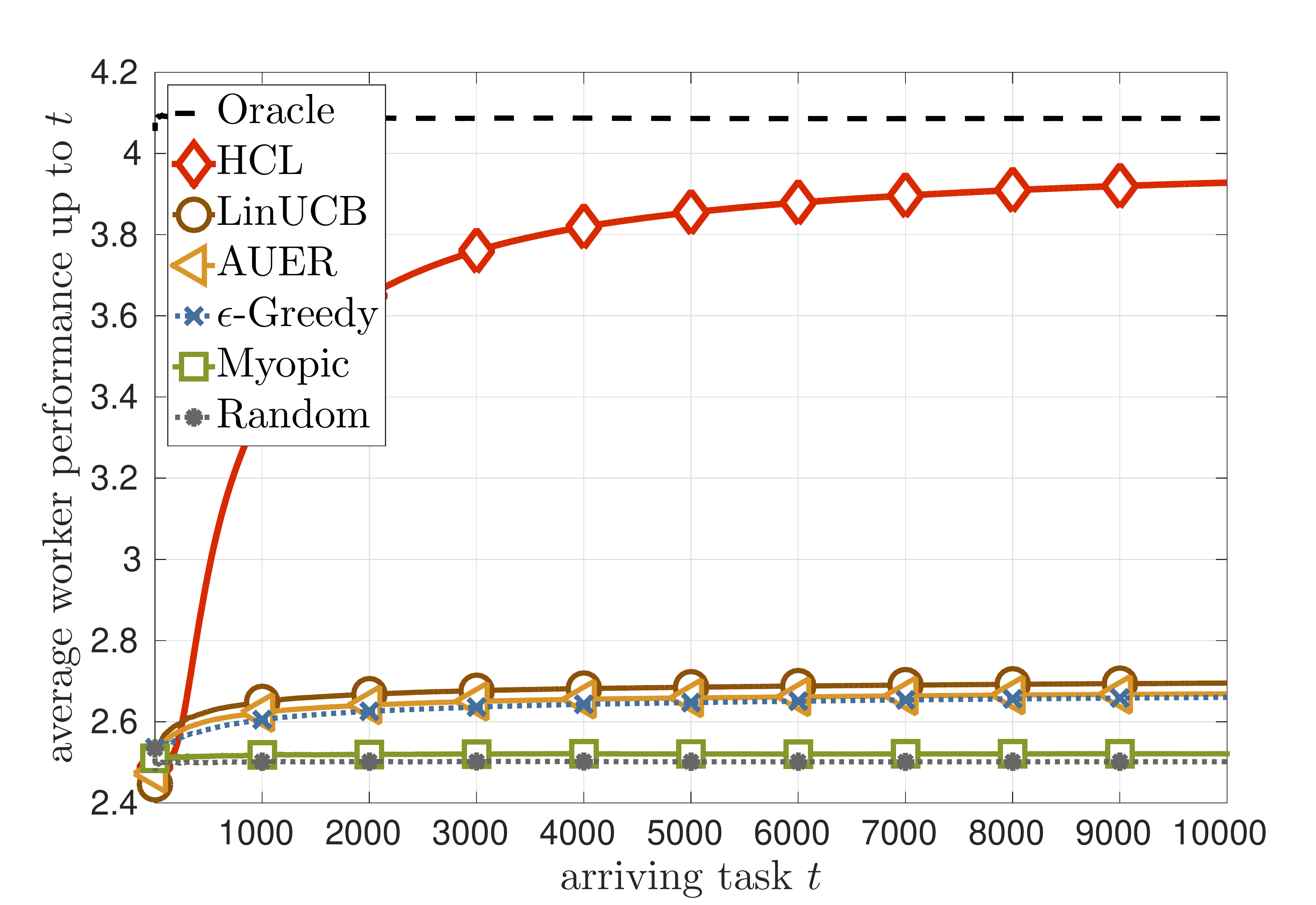}}
\hfil
\subfigure[Experiments with real data.]{\label{Fig_results_gow_av_perf_prav70}\includegraphics[width=0.46\textwidth]{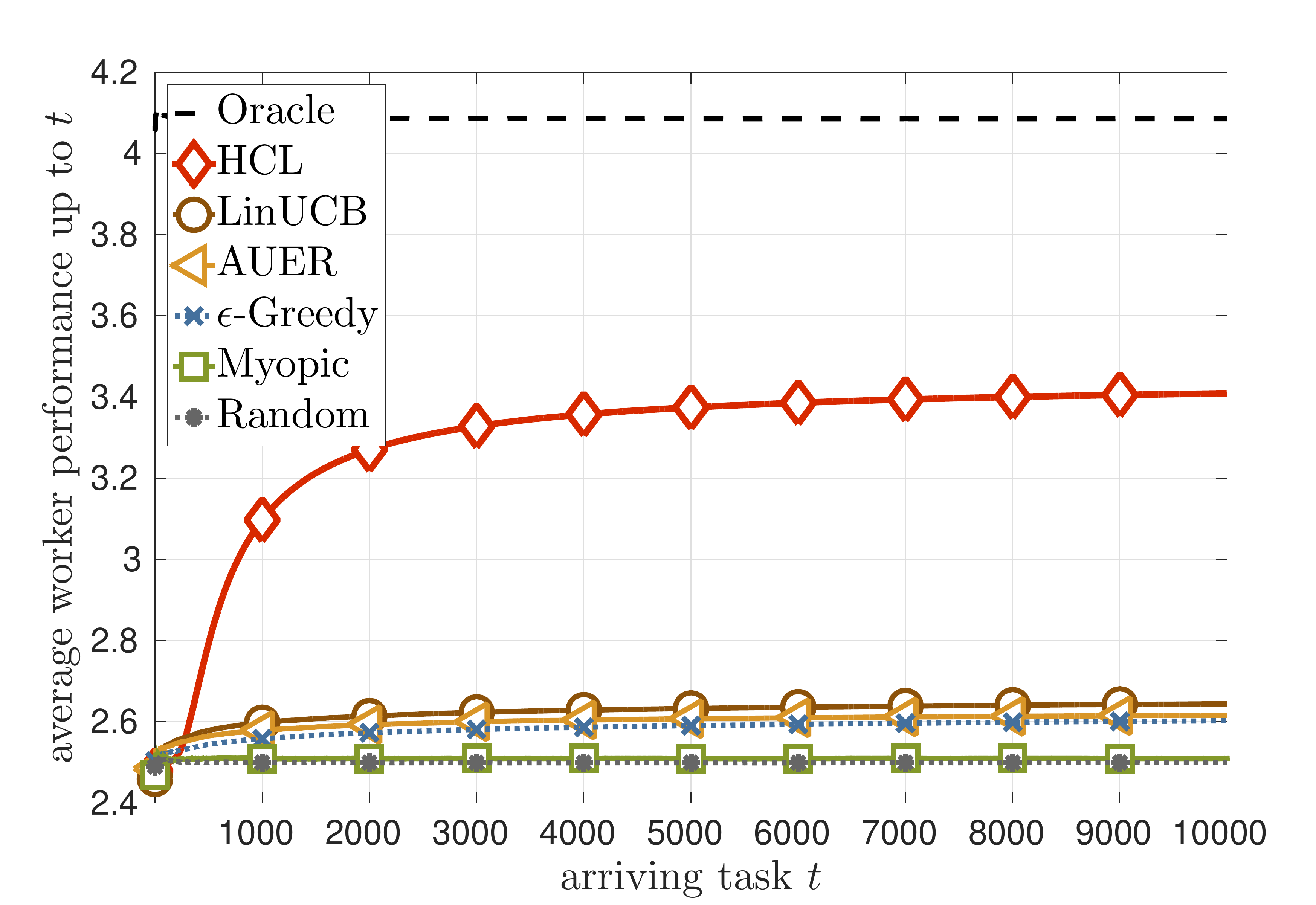}}
\caption{Average worker performance up to task $t$ for sequence $t=1,...,T$ for $\rho=0.7$ under the discrete performance model.}
\label{Fig_av_perf_prav70}
\end{figure}

Fig.~\ref{Fig_results_syn_av_perf_prav70}~and~Fig.~\ref{Fig_results_gow_av_perf_prav70} show the average worker performance up to task $t$ as a function of the sequentially arriving tasks $t=1,...,T$. 
We see that over the sequence of tasks, the average worker performance achieved by Oracle and Random stay nearly constant at around $4.1$ and $2.5$, respectively, for both synthetic and real data.
LinUCB, AUER, $\epsilon$-Greedy and Myopic increase the average worker performance slightly, starting between $2.4$ and $2.5$ at $t=1$ and ending with 
average performance of between $2.5$ and $2.7$ at $t=T$. 
On the contrary, $\PROPOSED$ is able to increase the average worker performance from $2.5$  at $t=1$ up to $3.9$ ($3.4$) at $t=T$ for the synthetic (real) data. 
Hence, $\PROPOSED$ learns context-specific worker performances and selects better workers over time.

\begin{figure}[!t]
\centering
\subfigure[Experiments with synthetic data.]{\label{Fig_syn_availability}\includegraphics[width=0.46\textwidth]{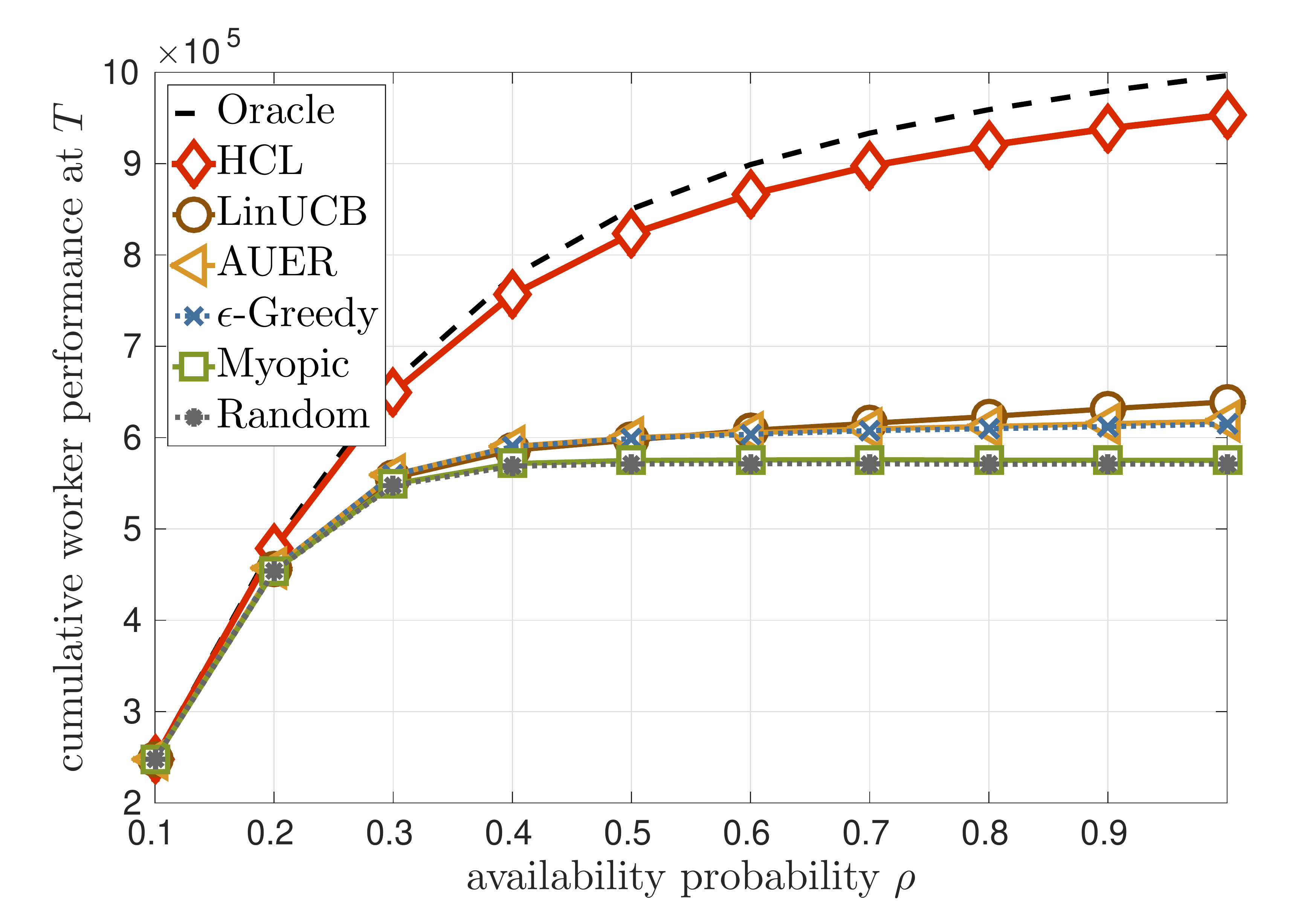}}
\hfil
\subfigure[Experiments with real data.]{\label{Fig_gow_availability}\includegraphics[width=0.46\textwidth]{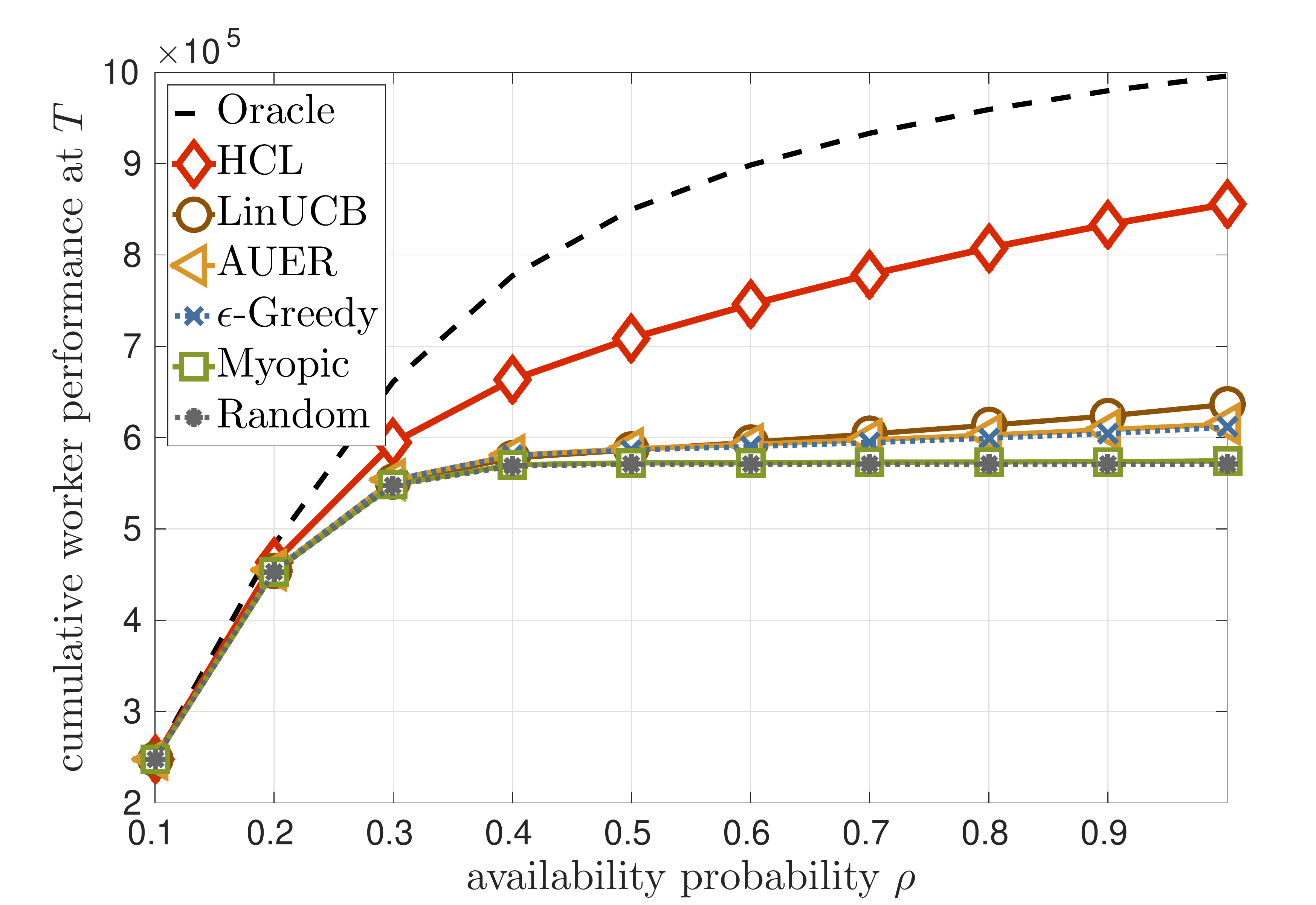}}
\caption{Impact of worker availability on cumulative worker performance at~$T$ for $T=10,000$ tasks under the discrete performance model.}
\label{Fig_availability}
\end{figure}

Finally, we evaluate the impact of worker availability by varying the parameter $\rho$. 
For each value of $\rho$, we average the results over $100$ synthetic instances and over $100$ real instances for $T=10,000$, respectively. 
Fig.~\ref{Fig_syn_availability}~and~\ref{Fig_gow_availability} show the cumulative worker performance at $T$ achieved by the algorithms for different~$\rho$. 
For small $\rho=0.1$, all algorithms yield approximately the same performance. This is as expected since given our modeling of task budget, for small~$\rho$, the number of available workers is often smaller than the required number of workers. Since each of the algorithm enters a select-all-workers phase in this case, each algorithm performs optimally. 
For increasing worker availability $\rho$, the cumulative performance at $T$ achieved by each of the algorithm increases. 
However, the gap between Oracle and $\PROPOSED$ on the one hand, and the remaining algorithms on the other hand, is increasing for increasing $\rho$. 
For example, at $\rho\in\{0.3,0.7,1\}$, the cumulative performance achieved by $\PROPOSED$ corresponds to $\{1.16, 1.46, 1.49\}$ ($\{1.07, 1.29, 1.34\}$) times the one achieved by the respective next best algorithm $\{$AUER, LinUCB, LinUCB$\}$ ($\{\epsilon$-Greedy, LinUCB, LinUCB$\}$) for the synthetic (real) data.
Hence, the more workers are available, the more severe is the effect of not selecting the best workers and only $\PROPOSED$ is able to cope with the more difficult worker selection.

\subsection{Results under the Hybrid Performance Model}\label{Subsec_HybridModel}
\begin{figure}[!t]
\centering
\includegraphics[width=0.46\textwidth]{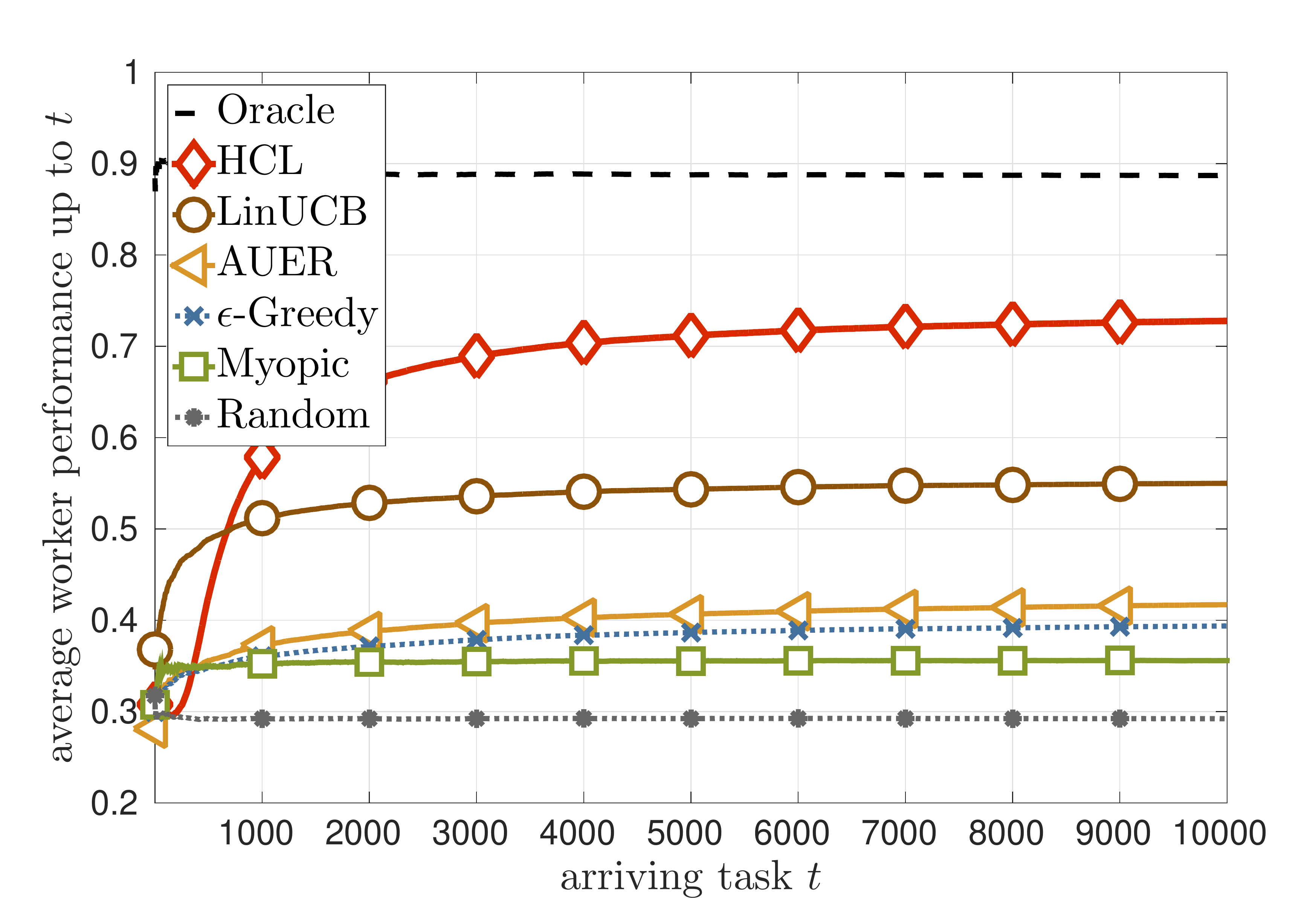}
\caption{Average worker performance up to task $t$ for sequence $t=1,...,T$ for $\rho=0.7$ under the hybrid performance model using real data.}
\label{Fig_av_perf_user_test_prav70}
\end{figure}

First, we run the algorithms on $100$ real instances for $T=10,000$ and $\rho=0.7$ using the hybrid performance model. Fig.~\ref{Fig_av_perf_user_test_prav70} shows the average worker performance up to task $t$ as a function of the sequentially arriving tasks $t=1,...,T$.\footnote{Note that worker performance is differently distributed in the hybrid than in the discrete model, so that the absolute values presented in Sec.~\ref{Subsec_HybridModel} are not comparable to those in Sec.~\ref{Subsec_DiscreteModel}.} The average worker performance achieved by Oracle and Random stay nearly constant at around $0.88$ and $0.29$ over the sequence of tasks.
AUER, $\epsilon$-Greedy and Myopic increase the average worker performance only slightly, from between $0.28$ and $0.31$ at $t=1$ to between $0.36$ and $0.42$ at $t=T$. 
LinUCB has a larger increase from $0.37$ at $t=1$ to $0.55$ at $t=T$. Compared to the discrete performance model, LinUCB performs better here due to the monotonic dependency of expected performance on battery state. 
Still, $\PROPOSED$ has the largest increase from $0.31$ at $t=1$ up to $0.73$ at $t=T$.
\begin{figure}[!t]
\centering
\includegraphics[width=0.46\textwidth]{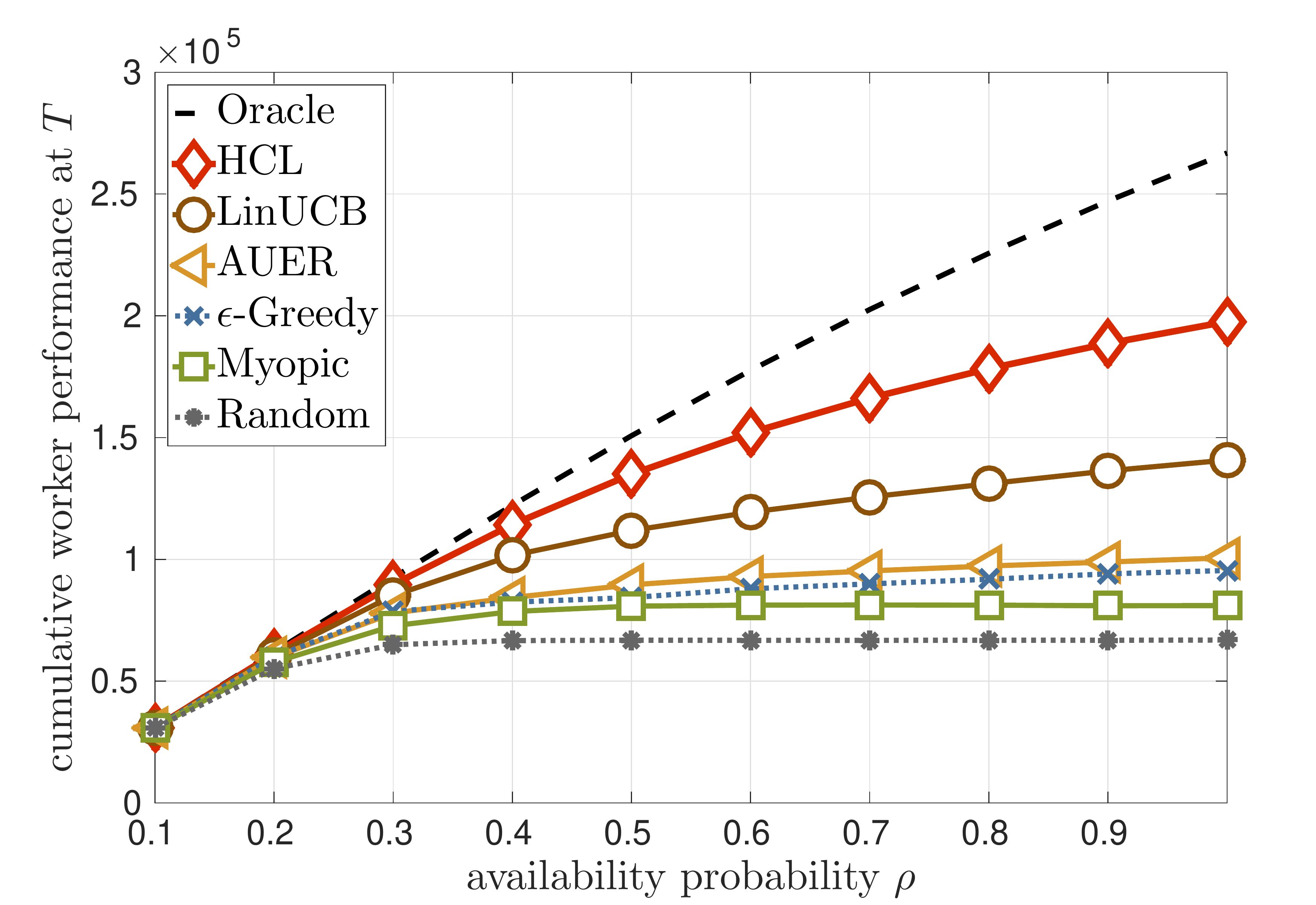}
\caption{Impact of worker availability on cumulative worker performance at $T$ for $T=10,000$ tasks under the hybrid performance model using real data.}
\label{Fig_availability_user_test}
\end{figure}
Finally, we evaluate the impact of worker availability~$\rho$.
For each value of $\rho$, we average the results over $100$ real instances for $T=10,000$.
Fig.~\ref{Fig_availability_user_test} shows the cumulative worker performance at~$T$ achieved by the algorithms for different~$\rho$. 
Again, for higher $\rho$, the algorithms achieve higher cumulative performances at $T$. 
While LinUCB performs better compared to the results under the discrete performance model, still, the gap in cumulative performance between $\PROPOSED$ and LinUCB is increasing for increasing $\rho$. 
For example, at $\rho\in\{0.3,0.7,1\}$, the cumulative performance achieved by $\PROPOSED$ corresponds to $\{1.05, 1.32, 1.40\}$ times the one achieved by LinUCB.

\section{Conclusion}\label{Sec_conclusion}
In this paper, we presented a context-aware hierarchical online learning algorithm, which learns context-specific worker performance online over time in order to maximize the performance in an MCS system for location-independent tasks. 
Our algorithm is split into two parts, one executed by LCs in the mobile devices of the workers, the other executed by the central MCSP. 
While the LCs learn their workers' performances, the MCSP assigns workers to tasks based on regular information exchange with the LCs.
Our hierarchical approach ensures that the most suitable workers are requested by the MCSP. 
The learning in LCs ensures that personal worker context can be kept locally, but still workers are offered those tasks they are interested in the most.
We showed that the requirements of our algorithm in terms of storage, communication and the number of quality assessments are small. Moreover, the algorithm converges to the optimal task assignment strategy.

\section*{Acknowledgement}
The work by S.~Klos and A.~Klein has been funded by the German Research Foundation (DFG) as part of project B3 within the Collaborative Research Center 1053 -- MAKI. 
The work by C. Tekin was supported by the Scientific and Technological
Research Council of Turkey (TUBITAK) under 3501 Program Grant
116E229.
The work by M.~van~der~Schaar was supported by an ONR Mathematical Data Sciences grant and by NSF~1524417 and NSF~1462245 grants.

\appendices
\section{Proof of Theorem 1}\label{App_regret}

Given any length $T$ sequence of task and worker arrivals, let $\tau_T$ be the set of tasks in $\{1,...,T\}$ for which $W_t > m_t$, and $\tau^c_T = \{1,...,T\} \setminus \tau_T$. 
$\tau^c_T$ is also called the set of select-all-workers phases. 
Also let $\tilde{\tau}_T\subseteq \tau_T$ be the set of tasks in $\tau_T$ for which the MCSP is in exploitation phase, and $\tilde{\tau}^c_T = \tau_T\setminus \tilde{\tau}_T$ be the set of tasks in $\tau_T$ for which the MCSP is in exploration phase. 
$\tilde{\tau}_T$ and $\tilde{\tau}^c_T$ are random sets that depend on the selections of the MCSP and the randomness of the observed performances. 
Using the expressions above, the regret can be decomposed as follows: 
\begin{align*}
R(T) &= \E\left[R_{\all}(T) + R_{\explore}(T) + R_{\exploit}(T) \right],
\end{align*}
where 
\begin{align*}
R_{\all}(T) &:= \sum_{t \in \tau^c_T} \sum_{j=1}^{\min\{m_t,W_t\}}\hspace{-0.2in} \bigl(\theta_{s_{t,j}^{*}}(x_{t,s_{t,j}^{*}},c_t) 
- \theta_{s_{t,j}}(x_{t,s_{t,j}},c_t)\bigr) \\ 
R_{\explore}(T) &:= \sum_{t \in \tilde{\tau}^c_T} \sum_{j=1}^{\min\{m_t,W_t\}} \hspace{-0.2in} \bigl(\theta_{s_{t,j}^{*}}(x_{t,s_{t,j}^{*}},c_t) 
- \theta_{s_{t,j}}(x_{t,s_{t,j}},c_t)\bigr) \\ 
R_{\exploit}(T) &:= \sum_{t \in \tilde{\tau}_T} \sum_{j=1}^{\min\{m_t,W_t\}} \hspace{-0.2in} \bigl(\theta_{s_{t,j}^{*}}(x_{t,s_{t,j}^{*}},c_t) 
- \theta_{s_{t,j}}(x_{t,s_{t,j}},c_t)\bigr) .
\end{align*}
Here, $R_{\all}(T)$, $R_{\explore}(T)$ and $R_{\exploit}(T)$ represent the regret due to select-all-workers phases, due to exploration phases and due to exploitation phases, respectively.
The regret is computed by considering for each task the loss due to selecting workers $\{s_{t,j}\}_{j=1,...,\min\{m_t,W_t\}}$ instead of the optimal workers $\{s_{t,j}^{\ast}\}_{j=1,...,\min\{m_t,W_t\}}$.
This loss is computed by subtracting the sum of expected performances of the optimal workers from the sum of expected performances of the selected workers.

Next, we will bound the expected values of each of the three summands above separately. 
First, we show that the regret due to select-all-workers phases is $0$.

\begin{Lemma}[Value of {$\E[R_{\all}(T)]$}]\label{Lemma_Ra}
When LC~$i \in {\cal W}$, runs Alg.~1 with an arbitrary deterministic function $K_i:\{1,...,T\}\rightarrow \mathbb{R}_{+}$ and an arbitrary $h_{T,i}\in \mathbb{N}$, and the MCSP runs Alg.~2, the regret $\E\left[R_{\all}(T)\right]$ satisfies
\begin{align}
 \E[R_{\all}(T)]  = 0.
\end{align}
\end{Lemma}
\begin{IEEEproof}[Proof of Lemma \ref{Lemma_Ra}]
For $t\in \tau^c_T$, i.e., $W_t\leq m_t$, the MCSP enters a select-all-workers phase. 
Moreover, for $W_t\leq m_t$, the trivial optimal solution is to request all available workers to complete task~$t$. 
Hence, the MCSP's selection of workers is optimal and therefore, select-all-workers phases do not contribute to the regret, i.e., $\E[R_{\all}(T)]  = 0$.
\end{IEEEproof}

Next, a bound for $\E\left[R_{\explore}(T)\right]$ is given. 

\begin{Lemma}[Bound for {$\E[R_{\explore}(T)]$}]\label{Lemma_Ror}
When LC $i$, $i \in {\cal W}$, runs Alg.~1 with parameters $K_i(t) = t^{z_i} \log (t)$, $t=1,...,T$, and $h_{T,i} = \ceil{T^{\gamma_i}}$, where $0<z_i<1$ and $0<\gamma_i <\frac{1}{D_i}$, 
and the MCSP runs Alg.~2, the regret $\E\left[R_{\explore}(T)\right]$ is bounded by
\begin{align}
 \E[R_{\explore}(T)]  &\leq W q_{\max} \sum_{i\in\mathcal{W}} 2^{D_i} (\log(T) T^{z_i + \gamma_i D_i} + T^{\gamma_i D_i}).
\end{align}
\end{Lemma}

\begin{IEEEproof}[Proof of Lemma \ref{Lemma_Ror}]
Let $t\in \tilde{\tau}^c_T$ be a task for which the MCSP enters an exploration phase. By design of $\PROPOSED$, in this case, it holds that $W_t> m_t$, i.e., $m_t=\min\{m_t,W_t\}$.
Since the expected performance of a worker is bounded in $[0,q_{\max}]$, it follows that
\begin{align*}
R_{\explore}(T) &= \sum_{t \in \tilde{\tau}^c_T}
\sum_{j=1}^{m_t} \left(\theta_{s_{t,j}^{*}}(x_{t,s_{t,j}^{*}},c_t) - \theta_{s_{t,j}}(x_{t,s_{t,j}},c_t)\right)\\
 &\leq \sum_{t \in \tilde{\tau}^c_T} m_t q_{\max}.
 \end{align*}
 Hence, the regret can be bounded by
 \begin{align*}
\E[R_{\explore}(T)] &\leq \E\left[\sum_{t \in \tilde{\tau}^c_T} m_t q_{\max}\right]\\
&\leq W q_{\max} \E\left[\sum_{t \in \tilde{\tau}^c_T} 1\right],
 \end{align*}
 since $m_t\leq W$ holds for all $t=1,...,T$.
 
For $t\in \tilde{\tau}^c_T$, the set of under-explored workers $\mathcal{W}_{t}^{\ue}$ is non-empty. Hence, there exists an available worker $i\in \mathcal{W}_t$ with 
$N_{i,q_{t,i}}(t)\leq K_i(t) = t^{z_i} \log (t)$. 
By definition of $\mathcal{W}_{t}^{\ue}$, up to task $T$, worker $i$ can induce at most $\ceil{T^{z_i} \log(T)}$ exploration phases for each of the $(h_{T,i})^{D_i}$ hypercubes of the partition $\mathcal{Q}_{T,i}$. 
Hence, the number of exploration phases is upper bounded as follows:
\begin{align*}\label{no_exploration}
\E\left[\sum_{t \in \tilde{\tau}^c_T} 1\right]\leq \sum_{i\in\mathcal{W}} (h_{T,i})^{D_i} \ceil{T^{z_i} \log(T)}.
\end{align*}
(This upper bound is rather loose, since several workers might be explored simultaneously, in which case they do not induce separate exploration phases.)
Hence, we conclude
\begin{align*}
 \E[R_{\explore}(T)] \leq W q_{\max}\sum_{i\in\mathcal{W}} (h_{T,i})^{D_i} \ceil{T^{z_i} \log(T)}.
\end{align*}
Using $(h_{T,i})^{D_i} = \ceil{T^{\gamma_i}}^{D_i}\leq (2T^{\gamma_i})^{D_i}= 2^{D_i} T^{\gamma_i D_i}$, it holds
\begin{align*}
 \E[R_{\explore}(T)]  &\leq W q_{\max} \sum_{i\in\mathcal{W}} 2^{D_i} (\log(T) T^{z_i + \gamma_i D_i} + T^{\gamma_i D_i}).
\end{align*}
\end{IEEEproof}

Next, we give a bound for $\E\left[R_{\exploit}(T)\right]$. 
\begin{Lemma}[Bound for {$\E\left[R_{\exploit}(T)\right]$}]\label{Lemma_Roi}
Given that Assumption~1 holds, when LC $i$, $i \in {\cal W}$, runs Alg.~1 with parameters $K_i(t) = t^{z_i} \log (t)$, $t=1,...,T$, and $h_{T,i} = \ceil{T^{\gamma_i}}$, where $0<z_i<1$ and $0<\gamma_i <\frac{1}{D_i}$, 
and the MCSP runs Alg.~2, the regret $\E\left[R_{\exploit}(T)\right]$ is bounded by
\begin{align}
\E[R_{\exploit}(T)] &\leq 2 \sum_{i\in\mathcal{W}} q_{\max}\frac{T^{1-\frac{z_i}{2}}}{1-\frac{z_i}{2}}+ 2 \sum_{i\in\mathcal{W}} L {D_{i}}^{\frac{\alpha}{2}}T^{1-\alpha \gamma_i}\notag\\
& \quad + q_{\max} W^2 \frac{\pi^2}{3}.
\end{align}
\end{Lemma}

\begin{IEEEproof}[Proof of Lemma \ref{Lemma_Roi}]
Let $t \in \tilde{\tau}_T$, i.e., the MCSP enters an exploitation phase. 
By design of $\PROPOSED$, in this case, it holds $W_t> m_t$, i.e., $m_t=\min\{m_t,W_t\}$. 
Additionally, since in exploitation phases, the set of under-explored workers is empty, i.e., $\mathcal{W}_{t}^{\ue}= \emptyset$, it holds that
$N_{i,q_{t,i}}(t)> K_i(t) = t^{z_i} \log (t)$ for all available workers $i\in \mathcal{W}_t$. 

Now, let $V(t)$ be the event that at the arrival of task $t$, each available worker $i$'s estimated performance $\hat{\theta}_{i,q_{t,i}}(t)$ in the current hypercube $q_{t,i}$ is ``close'' to its true expected value $\E[\hat{\theta}_{i,q_{t,i}}(t)]$, i.e.,
\begin{align*}
V(t) = \{|\hat{\theta}_{i,q_{t,i}}(t)-\E[\hat{\theta}_{i,q_{t,i}}(t)]|<H_i(t) \text{ for all } i\in \mathcal{W}_t\}
\end{align*}
for some arbitrary $H_i(t)>0$, $i\in \mathcal{W}_t$. 
Next, we distinguish between exploitation phases in which $V(t)$ or its complementary event, denoted by $V^c(t)$, hold. Let $I_{\{\cdot\}}$ denote the indicator function. Then, we can write
\begin{align*}
&R_{\exploit}(T) \\
&= \sum_{t \in \tilde{\tau}_T} \Biggl(I_{\{V(t)\}}\biggl( \sum_{j=1}^{m_t} (\theta_{ s_{t,j}^{\ast}}(x_{t,s_{t,j}^{\ast}}, c_t)- \theta_{ s_{t,j}}(x_{t,s_{t,j}}, c_t))\biggr)\Biggr)\\
&\quad \hspace{-0.2cm}+ \hspace{-0.1cm}\sum_{t \in \tilde{\tau}_T} \hspace{-0.1cm} \Biggl(I_{\{V^c(t)\}}\biggl( \sum_{j=1}^{m_t} (\theta_{ s_{t,j}^{\ast}}(x_{t,s_{t,j}^{\ast}}, c_t)- \theta_{ s_{t,j}}(x_{t,s_{t,j}}, c_t))\biggr)\hspace{-0.1cm}\Biggr).
\end{align*}
Using that the expected performance of a worker is bounded in $[0,q_{\max}]$, this term can further be bounded as
\begin{align}
&R_{\exploit}(T) \notag\\
&\leq \hspace{-0.1cm}\sum_{t \in \tilde{\tau}_T} \Biggl(I_{\{V(t)\}}\cdot\biggl( \sum_{j=1}^{m_t} (\theta_{ s_{t,j}^{\ast}}(x_{t,s_{t,j}^{\ast}}, c_t)- \theta_{ s_{t,j}}(x_{t,s_{t,j}}, c_t))\biggr)\hspace{-0.1cm}\Biggr)\label{Rs1}\\
&\quad + \sum_{t \in \tilde{\tau}_T} m_t q_{\max}I_{\{V^c(t)\}}.\label{E_Rs2}
\end{align}

First, we bound \eqref{Rs1}.
We start by noting that in an exploitation phase $t \in \tilde{\tau}_T$, since the MCSP selected workers $\{s_{t,j}\}_{j=1,...,m_t}$ instead of $\{s_{t,j}^{\ast}\}_{j=1,...,m_t}$, it holds that 
\begin{align}\label{Eq_opt_vs_select}
\sum_{j=1}^{m_t}\hat{\theta}_{ s_{t,j}^{\ast},q_{s_{t,j}^{\ast}}}(t)\leq \sum_{j=1}^{m_t}\hat{\theta}_{ s_{t,j},q_{s_{t,j}}}(t) .
\end{align}
We also know that when $V(t)$ holds, we have 
\begin{align}\label{Eq_V(t)}
\{|\hat{\theta}_{i,q_{t,i}}(t)-\E[\hat{\theta}_{i,q_{t,i}}(t)]|<H_i(t) \text{ for all } i\in \mathcal{W}_t\}
\end{align}
almost surely.
Finally, note that by H\"older continuity from Assumption~1, since $(x_{t,i},c_t) \in q_{t,i}$ and for calculating $\hat{\theta}_{i,q_{t,i}}(t)$, only contexts from hypercube $q_{t,i}$ are used, for each $i\in \mathcal{W}_t$, it follows that
\begin{align}\label{Ineq_partition}
&|\theta_{i}(x_{t,i},c_t) - \E[\hat{\theta}_{i,q_{t,i}}(t)]| \notag\\
&= \Biggl|\E\Biggl[\frac{1}{|\mathcal{E}_{i,q_{t,i}}(t)|} \sum_{p\in \mathcal{E}_{i,q_{t,i}}(t)} \bigl(\theta_{i}(x_{t,i},c_t) - p\bigr)\Biggr]\Biggr|\notag\\
&= \Biggl|\E\Biggl[\E\biggl[\frac{1}{|\mathcal{E}_{i,q_{t,i}}(t)|} \sum_{p\in \mathcal{E}_{i,q_{t,i}}(t)} \bigl(\theta_{i}(x_{t,i},c_t) - p\bigr)\biggl|\mathcal{E}_{i,q_{t,i}}(t) \biggr]\Biggr]\Biggr|\notag\\
&= \Biggl|\E\Biggl[\frac{1}{|\mathcal{E}_{i,q_{t,i}}(t)|} \sum_{p\in \mathcal{E}_{i,q_{t,i}}(t)}\biggl(\theta_{i}(x_{t,i},c_t) - \E\bigl[ p\bigl|\mathcal{E}_{i,q_{t,i}}(t) \bigr]\biggr)\Biggr]\Biggr|\notag\\
&\leq \E\Biggl[\frac{1}{|\mathcal{E}_{i,q_{t,i}}(t)|} \sum_{p\in \mathcal{E}_{i,q_{t,i}}(t)} L\Bigl|\Bigl|\bigr(\frac{1}{h_{T,i}}, \hdots ,\frac{1}{h_{T,i}}\bigr)\Bigr|\Bigr|^{\alpha}_i\notag\Biggr]\\
&\leq L {D_i}^{\frac{\alpha}{2}}h_{T,i}^{-\alpha},
\end{align}
where we used the definition of $\hat{\theta}_{i,q_{t,i}}(t)$ and the linearity of expectation in the first line, the law of total expectation in the second line and a property of conditional expectation in the third line. 
In the fourth line, we used triangle inequality and since the corresponding context of each of the observed performances $p\in \mathcal{E}_{i,q_{t,i}}(t)$ came from hypercube $q_{t,i}$, we used H\"older continuity from Assumption~1 and exploited the size $\frac{1}{h_{T,i}}\times \hdots \times\frac{1}{h_{T,i}}$ of the hypercubes.
Hence, by first using \eqref{Ineq_partition}, then \eqref{Eq_V(t)} and then \eqref{Eq_opt_vs_select}, we have for~\eqref{Rs1}
\begin{align}
& I_{\{ V(t)\}}
\cdot\Biggl( \sum_{j=1}^{m_t} \biggl(\theta_{ s_{t,j}^{\ast}}(x_{t,s_{t,j}^{\ast}}, c_t)- \theta_{ s_{t,j}}(x_{t,s_{t,j}}, c_t)\biggr)\Biggr) \notag \\
& \leq 
I_{\{ V(t)\}} \cdot
\Biggl(
\sum_{j=1}^{m_t} \biggl(\E[\hat{\theta}_{s_{t,j}^{\ast}, q_{s_{t,j}^{\ast}}}(t)]- \E[\hat{\theta}_{ s_{t,j}, q_{s_{t,j}}}(t)] \notag \\
&\quad 
+ L {D_{s^*_{t,j}}}^{\frac{\alpha}{2}}h_{T,{s^*_{t,j}}}^{-\alpha}
+ L {D_{s_{t,j}}}^{\frac{\alpha}{2}}h_{T,{s_{t,j}}}^{-\alpha}\biggr)
\Biggr) \notag \\
&\leq I_{\{ V(t)\}} \cdot
\Biggl( \sum_{j=1}^{m_t}\hat{\theta}_{ s_{t,j}^{\ast},q_{s_{t,j}^{\ast}}}(t) - \sum_{j=1}^{m_t}\hat{\theta}_{ s_{t,j},q_{s_{t,j}}}(t) \notag\\
&\quad + \sum_{j=1}^{m_t} \biggl(H_{s_{t,j}^{\ast}}(t) + H_{s_{t,j}}(t) \notag \\
&\quad  
+ L {D_{s^*_{t,j}}}^{\frac{\alpha}{2}}h_{T,{s^*_{t,j}}}^{-\alpha}
+ L {D_{s_{t,j}}}^{\frac{\alpha}{2}}h_{T,{s_{t,j}}}^{-\alpha}
\biggl)
\Biggl) \notag\\
&\leq \sum_{j=1}^{m_t} \biggl(H_{s_{t,j}^{\ast}}(t) + H_{s_{t,j}}(t) \notag \\
&\quad 
+ L {D_{s^*_{t,j}}}^{\frac{\alpha}{2}}h_{T,{s^*_{t,j}}}^{-\alpha}
+ L {D_{s_{t,j}}}^{\frac{\alpha}{2}}h_{T,{s_{t,j}}}^{-\alpha}
\biggr) \text{ almost surely.}\label{Rs1_2}
\end{align}
Taking the expectation of \eqref{Rs1} and exploiting that \eqref{Rs1_2} holds almost surely for any $t \in \tilde{\tau}_T$ yields
\begin{align*}
&\E[R_{\exploit}(T)] \notag\\
&\leq \sum_{t =1}^T \Biggl(\sum_{j=1}^{m_t} (H_{s_{t,j}^{\ast}}(t) + H_{s_{t,j}}(t) \notag \\
&\quad 
+ L {D_{s^*_{t,j}}}^{\frac{\alpha}{2}}h_{T,{s^*_{t,j}}}^{-\alpha}
+ L {D_{s_{t,j}}}^{\frac{\alpha}{2}}h_{T,{s_{t,j}}}^{-\alpha}
)\Biggr)\\
&\quad+ \E\left[\sum_{t \in \tilde{\tau}_T} m_t q_{\max}I_{\{V^c(t)\}}\right].
\end{align*}
Finally, adding non-negative summands and using $h_{T,i}^{-\alpha} = \ceil{T^{\gamma_i}}^{-\alpha}\leq T^{-\alpha \gamma_i}$, we further have
\begin{align*}
&\E[R_{\exploit}(T)] \notag\\
&\leq \sum_{t =1}^T \Biggl(2 \sum_{i\in\mathcal{W}} H_{i}(t) + 2 \sum_{i\in\mathcal{W}} L {D_{i}}^{\frac{\alpha}{2}}T^{-\alpha \gamma_i}\Biggr)\\
&\quad+ \E\left[\sum_{t \in \tilde{\tau}_T} m_t q_{\max}I_{\{V^c(t)\}}\right].
\end{align*}
Next, we take care of the term with the expected value in the last expression. It holds
\begin{align*}
&\E\left[\sum_{t \in \tilde{\tau}_T} m_t q_{\max}I_{\{V^c(t)\}}\right] \notag\\
&= \E\left[\E\biggl[\sum_{t \in \tilde{\tau}_T} m_t q_{\max}I_{\{V^c(t)\}}\biggl|\tilde{\tau}_T\biggr]\right] \notag\\
&= \E\left[\sum_{t \in \tilde{\tau}_T}m_t q_{\max}\E\biggl[ I_{\{V^c(t)\}}\biggl|\tilde{\tau}_T\biggr]\right] \notag\\
&= \E\left[\sum_{t \in \tilde{\tau}_T}m_t q_{\max}\Pr\bigl( V^c(t)\bigl|\tilde{\tau}_T\bigr)\right], \notag
\end{align*}
where we used the law of total expectation and a property of conditional expectation.

Next, we bound $\Pr(V^c(t)|\tilde{\tau}_T)$ for $t \in \tilde{\tau}_T$. 
The event $V^c(t)$ can be written as
\begin{align*}
V^c(t) = \{\exists i\in \mathcal{W}_t \text{ s.t. } |\hat{\theta}_{i,q_{t,i}}(t)-\E[\hat{\theta}_{i,q_{t,i}}(t)]|\geq H_i(t)\}.
\end{align*}
Hence,
\begin{align*}
& \Pr(V^c(t)|\tilde{\tau}_T) \\
&= \Pr(\exists i\in \mathcal{W}_t \text{ s.t. } |\hat{\theta}_{i,q_{t,i}}(t)-\E[\hat{\theta}_{i,q_{t,i}}(t)]|\geq H_i(t)|\tilde{\tau}_T)\\
&\leq \sum_{i\in \mathcal{W}_t} \Pr(|\hat{\theta}_{i,q_{t,i}}(t)-\E[\hat{\theta}_{i,q_{t,i}}(t)]|\geq H_i(t)|\tilde{\tau}_T).
\end{align*}
For $t \in \tilde{\tau}_T$, we get by the definition of $\mathcal{W}_{t}^{\ue}$, for each $i\in \mathcal{W}_t$ it holds that $N_{i,q_{t,i}}(t)> K_i(t) = t^{z_i} \log (t)$, and hence, $|\mathcal{E}_{i,q_{t,i}}(t)|>t^{z_i} \log (t)$.
For $i\in \mathcal{W}_t$ and $t \in \tilde{\tau}_T$, applying Hoeffding's inequality~\cite{Hoeffding1963} and using $|\mathcal{E}_{i,q_{t,i}}(t)|>t^{z_i} \log (t)$, we get 
\begin{align*}
&\Pr\bigl(|\hat{\theta}_{i,q_{t,i}}(t)-\E[\hat{\theta}_{i,q_{t,i}}(t)]|\geq H_i(t) | \tilde{\tau}_T\bigr)\\
&\leq 2\exp{\left(-2H_{i}(t)^2 t^{z_{i}} \log(t) \frac{1}{q_{\max}^2}\right)}.
\end{align*}

Hence, the regret due to exploitation phases is bounded by
\begin{align*}
& \E[R_{\exploit}(T)] \\
&\leq \sum_{t =1 }^T \Bigl(2 \sum_{i\in\mathcal{W}} H_{i}(t) + 2 \sum_{i\in\mathcal{W}} L {D_{i}}^{\frac{\alpha}{2}}T^{-\alpha \gamma_i}\Bigr)\\
&\quad  + \E\left[\sum_{t \in \tilde{\tau}_T}m_t q_{\max}\sum_{i\in\mathcal{W}_t} 2\exp{\left(-2H_{i}(t)^2 t^{z_{i}} \log(t) \frac{1}{q_{\max}^2}\right)}\right]\\
&\leq \sum_{t =1 }^T \Bigl(2 \sum_{i\in\mathcal{W}} H_{i}(t) + 2 \sum_{i\in\mathcal{W}} L {D_{i}}^{\frac{\alpha}{2}}T^{-\alpha \gamma_i}\Bigr)\\
&\quad + \sum_{t =1}^T m_t q_{\max}\sum_{i\in\mathcal{W}_t}2\exp{\left(-2H_{i}(t)^2 t^{z_{i}} \log(t) \frac{1}{q_{\max}^2}\right)}.
\end{align*}
So far, the analysis was performed with respect to some arbitrary $H_i(t)>0$, $i\in\mathcal{W}$. 
Setting $H_i(t) := q_{\max}t^{-\frac{z_i}{2}}$ for $i\in\mathcal{W}$, we get
\begin{align*}
& \E[R_{\exploit}(T)]\\
&\leq \sum_{t =1}^T \Bigl(2 \sum_{i\in\mathcal{W}} q_{\max}t^{-\frac{z_i}{2}} + 2 \sum_{i\in\mathcal{W}} L {D_{i}}^{\frac{\alpha}{2}}T^{-\alpha \gamma_i}\Bigr)\\
& \quad + \sum_{t=1 }^T m_t q_{\max}\sum_{i\in\mathcal{W}_t}2\exp{\left(\frac{-2q_{\max}^2 (t^{-\frac{z_i}{2}})^2 t^{z_{i}} \log(t)}{q_{\max}^2}\right)}\\
& \leq 2 \sum_{i\in\mathcal{W}} q_{\max}\sum_{t=1}^T t^{-\frac{z_i}{2}} + 2 \sum_{i\in\mathcal{W}} L {D_{i}}^{\frac{\alpha}{2}}T^{1-\alpha \gamma_i}\\
& \quad + q_{\max} W \sum_{t=1}^T m_t 2  t^{-2}\\
&\leq 2 \sum_{i\in\mathcal{W}} q_{\max}\frac{T^{1-\frac{z_i}{2}}}{1-\frac{z_i}{2}}+ 2 \sum_{i\in\mathcal{W}} L {D_{i}}^{\frac{\alpha}{2}}T^{1-\alpha \gamma_i}\\
&\quad + q_{\max} W^2 \frac{\pi^2}{3},
\end{align*}
where, in the last step, we used the result from Appendix~\ref{Appendix_Divergent_Series}, the fact that $m_t\leq W$ holds and the value of the Dirichlet series.
\end{IEEEproof}

Applying Lemmas \ref{Lemma_Ra}-\ref{Lemma_Roi}, the overall regret is bounded as follows. 

\begin{IEEEproof}[Proof of Theorem~1]
First, for $i\in\mathcal{W}$, let $K_i(t) = t^{z_i} \log (t)$ and $h_{T,i} = \ceil{T^{\gamma_i}}$, where $0<z_i<1$ and $0<\gamma_i <\frac{1}{D_i}$.
Then, under Assumption~1, by combining the results of Lemmas \ref{Lemma_Ra}-\ref{Lemma_Roi}, the regret $R(T)$ is bounded by
  \begin{align*}
 R(T) &\leq  q_{\max} W \sum_{i\in\mathcal{W}} 2^{D_i} (\log(T) T^{z_i + \gamma_i D_i} + T^{\gamma_i D_i}) \\
 & \quad+ 2 \sum_{i\in\mathcal{W}} q_{\max}\frac{T^{1-\frac{z_i}{2}}}{1-\frac{z_i}{2}}+ 2 \sum_{i\in\mathcal{W}} L {D_{i}}^{\frac{\alpha}{2}}T^{1-\alpha \gamma_i}\\
 & \quad+ q_{\max} W^2 \frac{\pi^2}{3}.
 \end{align*}
 The summands contribute to the regret with leading orders $O(\sum_{i\in\mathcal{W}} T^{z_i + \gamma_i D_i} \log(T))$, $O(\sum_{i\in\mathcal{W}} T^{1-\alpha \gamma_i})$ and $O(\sum_{i\in\mathcal{W}} T^{1-\frac{z_i}{2}})$.
 We balance the leading orders by setting the parameters $z_i, \gamma_i$ according to $z_i:= \frac{2\alpha}{3 \alpha + D_i}\in (0,1)$, 
 $\gamma_i := \frac{z_i}{2\alpha}\in (0,\frac{1}{D_i})$.
Then, the regret $R(T)$ is bounded by
   \begin{align*}
R(T) &\leq  q_{\max}W \sum_{i\in\mathcal{W}} 2^{D_i} (\log(T) T^{\frac{2\alpha + D_i}{3 \alpha + D_i}} + T^{\frac{D_i}{3 \alpha + D_i}}) \\
 & \quad+ \sum_{i\in\mathcal{W}} \frac{2q_{\max}}{(2\alpha + D_i)/(3 \alpha + D_i)}T^{\frac{2\alpha + D_i}{3 \alpha + D_i}}\\
 & \quad+  2 \sum_{i\in\mathcal{W}} L D_i^{\frac{\alpha}{2}}T^{\frac{2\alpha + D_i}{3 \alpha + D_i}} 
 + q_{\max} W^2 \frac{\pi^2}{3}.
 \end{align*}
 Setting $D_{\max}:=\max_{i\in\mathcal{W}}D_i$, the leading order of the regret is hence
$O\left(q_{\max}W^2 T^{\frac{2\alpha + D_{\max}}{3\alpha + D_{\max}}} \log(T) \right)$.
\end{IEEEproof}

\section{Proof of Corollary~1}\label{App_thm_qual_ass}
\begin{IEEEproof}[Proof of Corollary~1]
This follows directly from the proof of Lemma~\ref{Lemma_Ror} and the proof of Theorem~1. 
A quality assessment is only requested if a worker is selected for exploration purposes.
From the proof of Lemma~\ref{Lemma_Ror}, the number of times a worker can at most be selected for exploration purposes is upper bounded by $(h_{T,i})^{D_i} \ceil{T^{z_i} \log(T)} = \ceil{T^{\gamma_i}}^{D_i}\ceil{T^{z_i} \log(T)}\leq (1+T^{\gamma_i})^{D_i}(1+T^{z_i} \log(T))$.
Setting the parameters $z_i, \gamma_i$ as in the proof of Theorem~1 concludes the proof.
\end{IEEEproof}

\section{A BOUND ON DIVERGENT SERIES}\label{Appendix_Divergent_Series}
For $p>0$, $p\neq 1$, the following formula holds:
\begin{align*}
\sum_{t=1}^T \frac{1}{t^p} \leq 1 + \frac{T^{1-p}-1}{1-p}
\end{align*}
\begin{IEEEproof}
See \cite{Chlebus2009}.
\end{IEEEproof}


\end{document}